\documentclass{article}

\usepackage{microtype}
\usepackage{graphicx}
\usepackage{subfigure}
\usepackage{booktabs} 

\usepackage{hyperref}

\usepackage[accepted]{icml2022}

\usepackage{amsmath}
\usepackage{amssymb}
\usepackage[bb=dsserif]{mathalpha}
\usepackage{multirow}
\usepackage{multicol}
\usepackage{comment}
\allowdisplaybreaks

\newenvironment{proof}{\paragraph{Proof:}}{\hfill$\square$}
\newtheorem{myDef}{Definition} 
\newtheorem{myTh}{Theorem}

\icmltitlerunning{Improved Certified Defenses against Data Poisoning with (Deterministic) Finite Aggregation}

\begin{document}

\twocolumn[
\icmltitle{Improved Certified Defenses against Data Poisoning with\\ (Deterministic) Finite Aggregation}



\icmlsetsymbol{equal}{*}

\begin{icmlauthorlist}
\icmlauthor{Wenxiao Wang}{UMD}
\icmlauthor{Alexander Levine}{UMD}
\icmlauthor{Soheil Feizi}{UMD}
\end{icmlauthorlist}

\icmlaffiliation{UMD}{Department of Computer Science, University of Maryland, College Park, Maryland, USA}

\icmlcorrespondingauthor{Wenxiao Wang}{wwx@umd.edu}

\icmlkeywords{Machine Learning, ICML}

\vskip 0.3in
]



\printAffiliationsAndNotice{}  

\begin{abstract}
Data poisoning attacks aim at manipulating model behaviors through distorting training data. Previously, an aggregation-based certified defense, Deep Partition Aggregation (DPA), was proposed to mitigate this threat. 
DPA predicts through an aggregation of base classifiers trained on disjoint subsets of data, thus restricting its sensitivity to dataset distortions. 
In this work, we propose an improved certified defense against general poisoning attacks, namely \textbf{Finite Aggregation}. 
In contrast to DPA, which directly splits the training set into {\it disjoint} subsets, our method first splits the training set into smaller disjoint subsets and then combines duplicates of them to build larger (but not disjoint) subsets for training base classifiers. 
This reduces the worst-case impacts of poison samples and thus improves certified robustness bounds. 
In addition, we offer an alternative view of our method, bridging the designs of deterministic and stochastic aggregation-based certified defenses. Empirically, our proposed Finite Aggregation consistently improves certificates on MNIST, CIFAR-10, and GTSRB, boosting certified fractions by up to $3.05\%$, $3.87\%$ and $4.77\%$, respectively, while keeping the same clean accuracies as DPA's, effectively establishing a new \textbf{state of the art} in (pointwise) certified robustness against data poisoning.

\end{abstract}

\begin{figure*}[!tbp]
\begin{center}
\includegraphics[width=\linewidth]{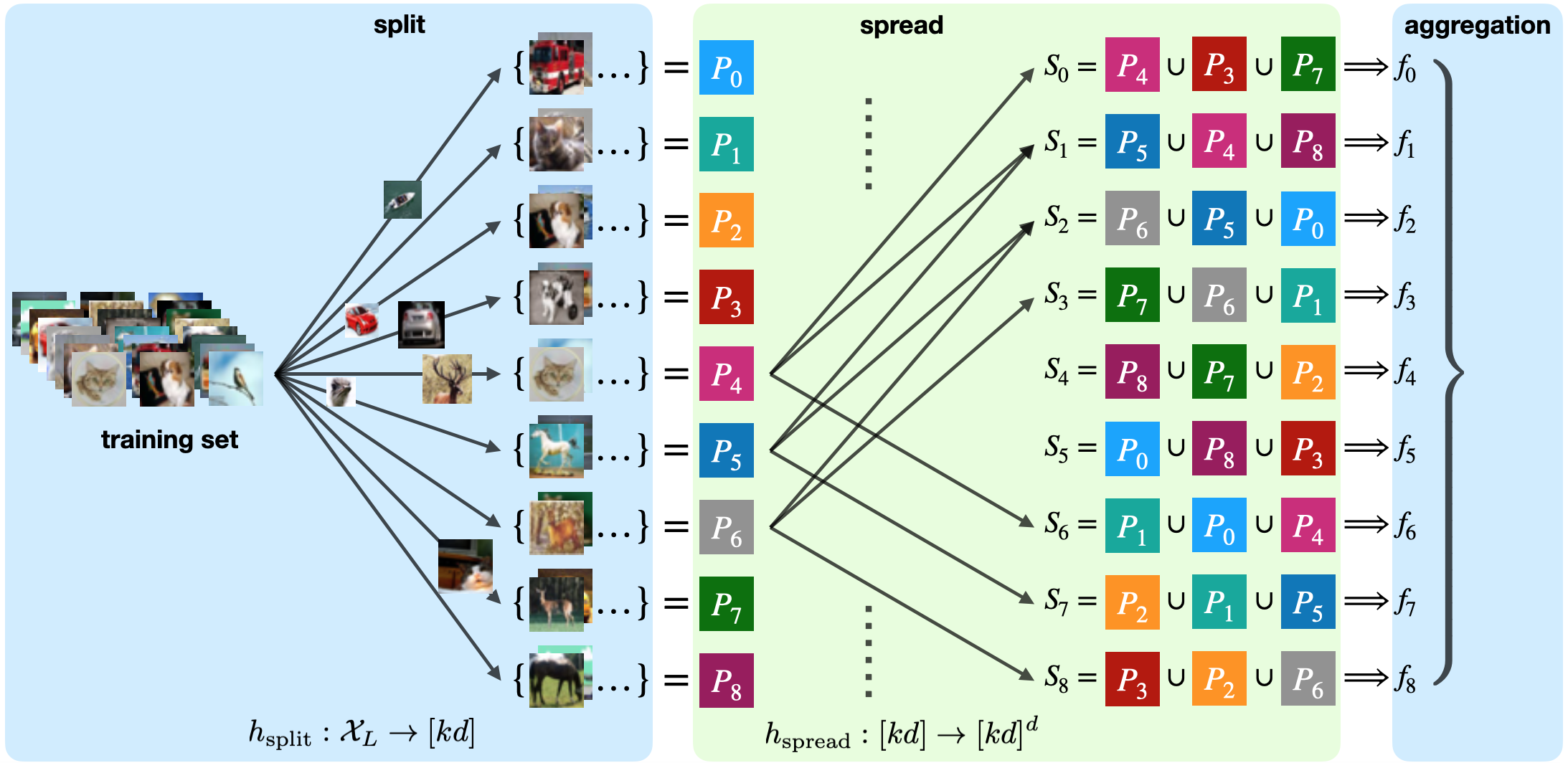}
\caption{An overview of \textbf{Finite Aggregation} with $k=3$ and $d=3$. Finite Aggregation consists of three parts: \textbf{split}, where the training set is split into $kd$ partitions $P_0, \dots, P_{kd-1}$ using a hashing function $h_\text{split}$; \textbf{spread}, where each partition is spread, according to a hash function $h_\text{spread}$, to $d$ different destinations from a total of $kd$ subsets $S_0, \dots, S_{kd-1}$; and \textbf{aggregation}, where one classifier is trained from every subset and the majority vote of all $kd$ classifiers will be the prediction at inference time.  }
\label{fig:FA}
\end{center}
\end{figure*}

\begin{figure}[!tbp]
\begin{center}
\includegraphics[width=\linewidth]{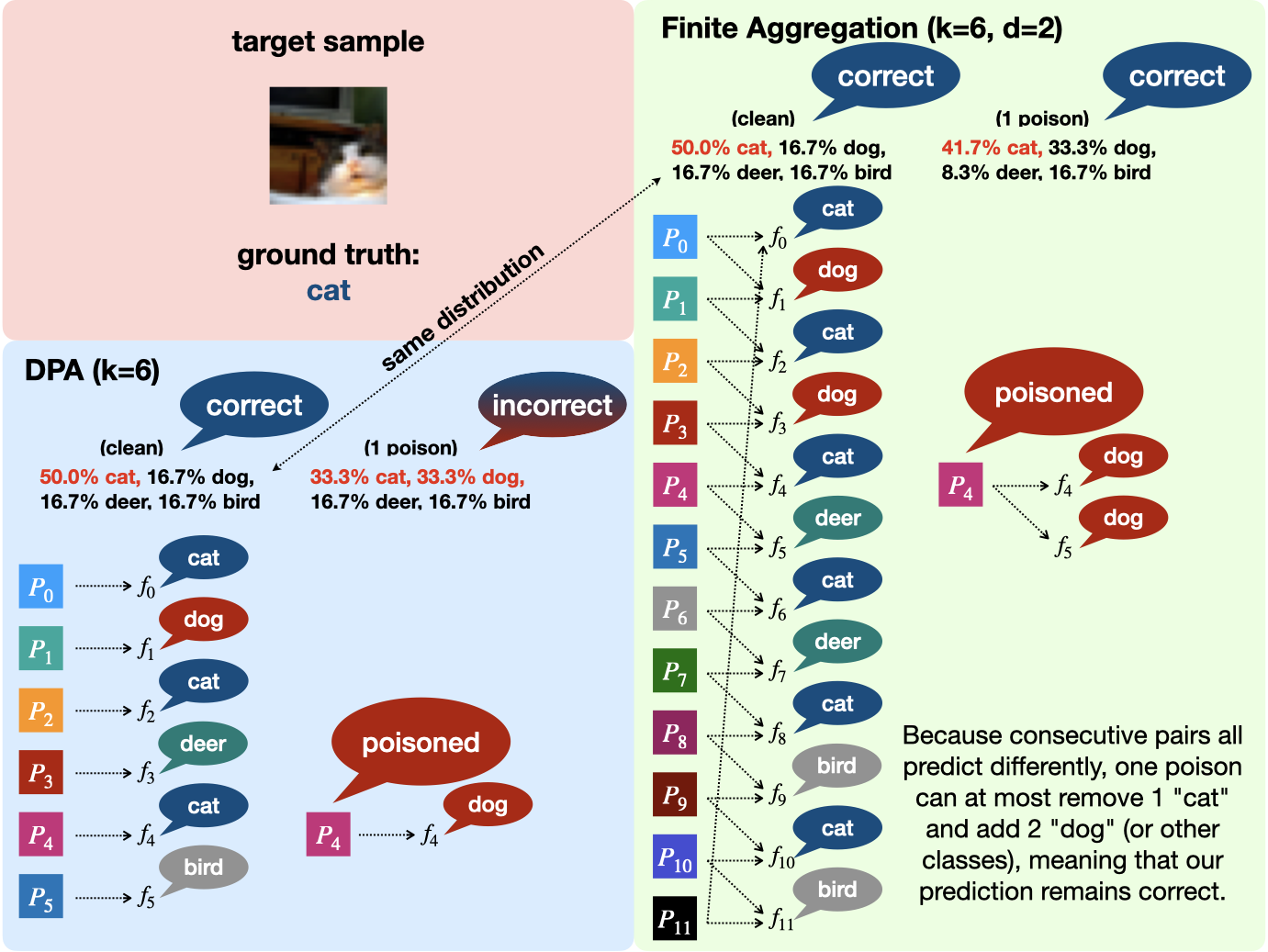}
\caption{A toy example illustrating how our proposed Finite Aggregation improves provable robustness through a strategic reusing of every sample. 
For simplicity, we assume that every partition in this example contributes to two consecutive base classifiers in Finite Aggregation. 
Notably, with no poison, the distribution of predictions from base classifiers are identical for DPA and our method. 
However, since the subset of base classifiers corresponding to every sample may predict differently, a poisoned sample can be less effective in our method compared to in DPA, leading to improved robustness. }
\label{fig:FA_toy}
\end{center}
\end{figure}

\section{Introduction}
\label{sec:intro}

Over the past years, we have witnessed the increasing popularity of deep learning in a variety of domains including computer vision \cite{he2016deep}, natural language processing \cite{bert}, and speech recognition \cite{SpeechRecog}.
In many cases, such rapid developments depend heavily on the increased availability of data collected from diverse sources, which can be different users or simply websites from all over the Internet.
While the richness of data sources greatly facilitates the advancement of deep learning techniques and their applications, it also raises concerns about their {\it reliability}. This makes the data poisoning threat model, which concerns the reliability of models under adversarially corrupted training samples, more important than ever \cite{DatasetSecurity}.

In this work, we use a general formulation of data poisoning attacks as follows: The adversary is given the ability to insert/remove a bounded number of training samples in order to manipulate the predictions (on some target samples) of the model trained from the corresponding training set. 
Here, the number of samples that the adversary is allowed to insert/remove is referred to as the attack size.

Many variants of empirical poisoning attacks targeting deep neural networks have been proposed, including Feature Collision \cite{FeatureCollision}, Convex Polytope \cite{ConvexPolytope}, Bullseye Polytope \cite{Bullseye} and Witches' Brew \cite{Brew}. 
These attacks are also referred to as triggerless attacks since no modification to the targets is required. 
Unlike triggerless attacks, backdoor attacks are poisoning attacks that allow modifications of the target samples, for which a variety of approaches have been developed including backdoor poisoning \cite{TargetedBackdoor}, label-consistent backdooring \cite{LC_backdoor} and hidden-trigger backdooring \cite{HT_backdoor}. While it is shown in \cite{HowToxic} that the evaluation settings can greatly affect the success rate of many data poisoning attacks to deep models, the vulnerability issues against poisoning attacks remain because (i) the current attacks can still succeed in many scenarios, and (ii) stronger adaptive poisoning attacks can potentially be developed in the future, posing practical threats.

In this work, we focus on developing {\it provably} robust defenses against general poisoning attacks. 
In particular, aggregation-based techniques, including a deterministic one \cite{DPA} and stochastic ones \cite{Bagging, RandomSelection}, have been adopted to offer (pointwise) certified robustness against poisoning attacks, where the prediction on every sample is guaranteed to remain unchanged within a certain attack size. Notably, to date, they are state-of-the-art in providing (pointwise) certified robustness against general poisoning attacks.
We have other certified defenses against poisoning attacks discussed in Section \ref{sec:related_work}.

In this work, we present \textbf{Finite Aggregation}, an advanced aggregation-based defense extended from Deep Partition Aggregation (DPA) \cite{DPA}. 
DPA predicts through an aggregation of base classifiers trained on disjoint subsets of data, thus restricting its sensitivity to dataset distortions. 
While DPA simply splits the training set into disjoint subsets for training base classifiers, we introduce a novel `split\&spread' protocol to obtain more {\it overlapping} subsets without changing the average subset size, as illustrated in Figure \ref{fig:FA}:
We first split the training set into smaller disjoint subsets and then combine duplicates of them in a structured fashion to build larger (but \textbf{not disjoint}) subsets.

The key idea of our proposed {\it Finite Aggregation} is based on a strategic \textbf{reusing} of every sample to improve robustness. To certify a prediction against data poisoning, an implicit or explicit characterization of the worst-case impact of poisoned samples is inevitable, which can be made more fine-grained by reusing samples strategically to better assess the effectiveness of potentially poisoned samples. In particular, in our proposed Finite Aggregation method (Figure \ref{fig:FA}), since the subset of base classifiers corresponding to every sample may predict differently, a poisoned sample can be less effective compared to that of DPA, leading to improved robustness. We further illustrate this using a toy example in Figure \ref{fig:FA_toy}. In this example, having even one poison sample in DPA can create a tie between prediction probabilities of the correct class (`cat') and an incorrect class (`dog'). However, in Finite Aggregation, with the same clean distribution, even the most effective poison is not able to mislead the model (detailed in Appendix \ref{sec:toy_example_explain}).

One should note that in the Finite Aggregation method, the total number of base classifiers increases with the reusing of samples, which is why allowing every sample to be used by more base classifiers can actually reduce the impact of poisoned samples. Notably, as $d$ controls the sample reusing in Finite Aggregation, our method essentially degenerates to DPA when $d=1$ (i.e. no sample reusing).

In summary, our contributions in this work are as follows:
\begin{itemize}
\item We propose \textbf{Finite Aggregation}, an advanced aggregation-based provable defense against general data poisoning that obtains improved (pointwise) robustness bounds through strategic \textbf{reusing} of samples.

\item We offer \textbf{a novel, alternative view} of our design, to bridge the gap between deterministic defenses (e.g. \cite{DPA}) and stochastic aggregation-based defenses (e.g. \cite{Bagging, RandomSelection}).

\item Empirically, our method effectively improves certified fractions by up to $3.05\%$, $3.87\%$ and $4.77\%$ respectively on MNIST, CIFAR-10, and GTSRB, while keeping the same clean accuracies as DPA's, establishing a new \textbf{state of the art} in (pointwise) certified robustness against general data poisoning.
\end{itemize}

\section{Related Work}
\label{sec:related_work}

\textbf{Certified Robustness against Data Poisoning.}
While in this work we consider pointwise certified robustness, some prior works provide distributional robustness against data poisoning. To name a few, \cite{Cert_Def_poison} derives a high-probability lower bound for test accuracy under poisoning attacks, assuming the distribution of the testing set is the same as the one for the clean training set; \cite{DiakonikolasKKL16, LaiRV16} provides distributional robustness guarantees for certain types of unsupervised learning; \cite{Sever} offers provable approximations of the clean model with additional assumptions regarding the distribution of the clean training data. 
In addition, \cite{PAC_learn_cert_poisoning} studies conditions for learnability and certification on predictions under poisoning attacks through the scope of PAC learning; \cite{backdoor_RS,backdoor_RAB} study certified robustness against backdoor attacks; \cite{LabelFlipping_RS} studies certified robustness against label-flipping attacks.

\textbf{Privacy Attacks.} Privacy attacks and data poisoning attacks are actually related. While data poisoning attacks focus on the interests of the consumers of data (e.g. model trainers), privacy attacks focus on the interests of the providers of data (e.g. users). Intuitively, the idea of defending both attacks can be to have models that are not sensitive to the change of a single sample. 
An active field in privacy-preserving machine learning is to combine deep learning with differential privacy \cite{DPDL, PATE, PATE2, dplis}. Some existing works have related differential privacy with poisoning attacks: For example, \cite{DP_detect} proposes an approach of backdoor attack detection via differential privacy and \cite{Poison_DP} investigate the empirical effectiveness of differential privacy in defending against poisoning attacks.
\section{(Deterministic) Finite Aggregation}
\label{sec:method}

\subsection{Notation and Background}
\label{sec:notation_background}
Our design, \textbf{Finite Aggregation}, is extended from the framework of Deep Partition Aggregation (i.e., DPA) \cite{DPA}. 
In this section, we will go through the notations and the main results of DPA. 

\textbf{Notation:} 
Let $\mathcal{X}$ be the space of unlabeled samples (e.g. the space of images), $\mathcal{C}$ be the set of class indices $[n_c] = \{0, 1, \dots, n_c-1\}$, 
and $\mathcal{X}_{L}$ be the space of labeled samples $\{(x, c) | x\in \mathcal{X}, c\in \mathcal{C}\}$. 
A training set $D$ of size $n$ can be viewed as a multiset $\{(x_i, c_i)\}_{i=1}^n$ where $(x_i, c_i)\in \mathcal{X}_{L}$. 
We use $\mathcal{D}$ to denote the space of training sets. 

For a deterministic classification algorithm $f: \mathcal{D} \times \mathcal{X} \to \mathcal{C}$,  $f(D, x)\in \mathcal{C}$ denotes the predicted class index for input $x\in \mathcal{X}$ when the training set is $D \in \mathcal{D}$.

For two training sets $D$ and $D'$, we measure their difference with symmetric distance (the cardinality of symmetric difference): 
\begin{align*}
d_\text{sym}(D, D') = |(D\setminus D')\cup (D'\setminus D)|,
\end{align*}
which is exactly the minimum number of samples one needs to insert/remove to change one training set to another (i.e. change $D$ to $D'$ or change $D'$ to $D$).

\textbf{DPA \cite{DPA}:} DPA is a deterministic classification method $\text{DPA}: \mathcal{D} \times \mathcal{X} \to \mathcal{C}$ constructed using a deterministic base classifier $f_\text{base}: \mathcal{D} \times \mathcal{X} \to \mathcal{C}$ 
and a hash function $h: \mathcal{X}_L \to [k]$ that maps labeled samples to integers between $0$ and $k-1$ ($k$ is a hyperparameter denoting the number of partitions).
The construction is:
\begin{align*}
    \text{DPA} (D, x) = \arg \max_{c\in \mathcal{C}} \sum_{i=0}^{k-1} \mathbb{1}\left[  f_\text{base} (P_i, x) = c \right],
\end{align*}
where $P_i = \{ (x,c)\in D | h((x,c)) = i  \}$ is a partition containing all training samples with a hash value of $i$. Ties are broken by returning the smaller class index in $\arg \max$. For convenience, we use $\text{DPA}(D, x)_c$ to denote the average votes count $\frac{1}{k} \sum_{i=0}^{k-1} \mathbb{1}\left[  f_\text{base} (P_i, x) = c \right]$.

\begin{myTh}[Robustness of DPA against Data Poisoning]
\label{Th:DPA}
Given a training set $D$ and an input $x$, let $c = \text{DPA}(D, x)$, then for any training set $D'$, if 
\begin{align*}
    \frac{2}{k} \cdot d_\text{sym} (D, D') \leq \text{DPA}(D, x)_c - \text{DPA}(D, x)_{c'} - \frac{\mathbb{1}\left[c'<c\right]}{k}
\end{align*}
holds for all $c' \neq c$, we have $\text{DPA}(D, x) = \text{DPA}(D', x)$.
\end{myTh}

Theorem \ref{Th:DPA}\footnote{For coherence, Theorem \ref{Th:DPA} is presented in a slightly different form from the original one by \cite{DPA}.} \cite{DPA} shows how DPA offers certified robustness against data poisoning attacks. Intuitively, since every sample will be contained in only one partition, one poisoned sample can change at most one vote and therefore reduce normalized margins (the gap between average vote counts) by $\frac{2}{k}$ at most. Thus, the adversary can never change the prediction as long as the number of samples inserted/removed is limited (i.e. $d_\text{sym}$ is small).

\subsection{Proposed Design}
\label{sec:proposed_design}
In this section, we will present the design of our method, Finite Aggregation. \textbf{Finite Aggregation} constructs a new, deterministic classifier using the followings:
\begin{itemize}
    \item a deterministic base classifier $f_\text{base}: \mathcal{D} \times \mathcal{X} \to \mathcal{C}$;
    \item a hash function $h_\text{split}: \mathcal{X}_L \to [kd]$ mapping labeled samples to partition indices between $0$ and $kd-1$, which is used to split the training set into $kd$ partitions;
    \item a balanced hash function $h_\text{spread}: [kd] \to [kd]^d$ mapping every partition index to a set of $d$ different integers of the same range, which is used to spread training samples, allowing them to be utilized by $d$ different base classifiers. 
\end{itemize}
Here, $k$ and $d$ are two hyperparameters where $k$ corresponds to the inverse of sensitivity 
and the spreading degree $d$ controls the number of base classifiers that every sample can be utilized by.

By a balanced hash function, we mean that the inverse of the hash function ${h_\text{spread}^{-1}(i) = \{ j\in [kd] | i\in h_\text{spread}(j) \}}$ has the same size (i.e., $d$ elements) for all $i \in [kd]$, which means $h_\text{spread}^{-1}$ is also a hash function from $[kd]$ to $[kd]^d$. Our choice of $h_\text{spread}$ will be discussed in Section \ref{sec:PracDetail}.

\begin{myDef}[Classification with Finite Aggregation]
\label{Def:FA}
The construction of Finite Aggregation $\text{FA}: \mathcal{D} \times \mathcal{X} \to \mathcal{C}$ is as follows:
\begin{align*}
    &\text{FA}(D, x) = \arg \max_{c \in \mathcal{C}} \sum_{i=0}^{kd-1} \mathbb{1} \left[ f_\text{base}(S_i, x) = c \right],
\end{align*}
where $S_i = \bigcup_{j \in h_\text{spread}^{-1} (i)} P_j $,  $P_j = \{ (x,c)\in D | h_\text{split}((x,c)) = j\}$ and ties are broken by returning the smaller class index in $\arg \max$.
\end{myDef}

Similarly, we use $\text{FA}(D, x)_c$ to denote the the average votes count $\frac{1}{kd} \sum_{i=0}^{kd-1} \mathbb{1} \left[ f_\text{base}(S_i, x) = c \right]$.
We use $\text{FA}(D, x)_{c|j}$ to denote $\frac{1}{d}\sum_{i \in h_\text{spread}(j)} 
\mathbb{1} \left[ f_\text{base}(S_i, x) = c \right]$, which is the average votes count over base classifiers that utilize $P_j$.

An overview of Finite Aggregation is in Figure \ref{fig:FA}. 
Finite Aggregation can be decomposed into three stages:
\begin{itemize}
    \item \textbf{Split}, where the training set is split into $kd$ partitions $P_0, \dots, P_{kd-1}$;
    \item \textbf{Spread}, where each partition is spread to $d$ different destinations in $S_0, \dots, S_{kd-1}$; 
    \item \textbf{Aggregation}, where one classifier is trained from every subset $S_i$, $i\in[kd]$ and the majority vote of all $kd$ classifiers will be the prediction at inference time.
\end{itemize}

In both DPA \cite{DPA} and our design (with the same hyperparameter $k$), every base classifier will, on average, have access to $1/k$ of the entire training, and every sample will be utilized by exactly $1/k$ of the base classifiers. 
However, unlike DPA, which forms disjoint subsets, we let every sample be utilized by $d$ base classifiers in a way that enables better certifications against data poisoning attacks. Notably, when $d = 1$, Finite Aggregation essentially reduces to DPA with the same hyperparameter $\mathbf{k}$. 

\subsection{Certified Robustness to Data Poisoning}
\label{sec:robustness}
In this section, we will see how Finite Aggregation provably defends against data poisoning attacks and discuss why it offers a stronger defense than DPA.

\begin{myTh}[Finite Aggregation against Data Poisoning]
\label{Th:FA}
Given a training set $D$ and an input $x$, let $c = \text{FA}(D, x)$, then for any training set $D'$, it is guaranteed that $\text{FA}(D', x) = \text{FA}(D, x)$ when
\begin{align*}
    \frac{1}{k} \cdot \Delta_{D, x}^{\overline{d_\text{sym} (D, D')}}  \leq \text{FA}(D, x)_c - \text{FA}(D, x)_{c'} - \frac{\mathbb{1}\left[c'<c\right]}{kd}
\end{align*}
holds for all $c' \neq c$, where $\Delta_{D, x}$ is a multiset defined as 
\begin{align*}
\left\{ 1 + \text{FA}(D, x)_{c|j} - \text{FA}(D, x)_{c'|j} \right\}_{j \in [kd]}
\end{align*}
and $\Delta_{D, x}^{\overline{d_\text{sym} (D, D')}}$ denotes the sum of the largest $d_\text{sym} (D, D')$ elements in the multiset $\Delta_{D, x}$.
\end{myTh}

Theorem \ref{Th:FA} is how Finite Aggregation offers certified robustness against data poisoning. The detailed proof is in Appendix \ref{Pf:FA} and we include a sketch here.

When one sample $x'$ is inserted or removed, only a specific set of $d$ base classifiers may be affected, depending on where this sample is assigned to in the split stage (i.e. the value of $h_\text{split}(x')$). 
If the goal is to change the prediction from $c$ to $c'$, then the worst case is simply that all of those $d$ classifiers will predict $c'$ after the insertion/removal, meaning that the contribution to the margin between class $c$ and $c'$ will reduce by $\frac{2}{kd}$ for every base classifier (among those $d$) that originally predicts $c$, by $0$ for every that predicts $c'$ and by $\frac{1}{kd}$ for every base classifier that predicts other classes. 
Thus, the margin will be reduced by at most $\frac{1}{k} \left(1 + \text{FA}(D, x)_{c|j} - \text{FA}(D, x)_{c'|j} \right)$ given $h_\text{split}(x')=j$ 
and therefore with $d_\text{sym} (D, D')$ samples inserted/removed, the margin will be reduced by $\frac{1}{k}\Delta_{D, x}^{\overline{d_\text{sym} (D, D')}}$ at most, which means the prediction will not be turned from $c$ to $c'$ as long as the margin is larger than this. 

Comparing Theorem \ref{Th:FA} with the certified robustness of DPA in Theorem \ref{Th:DPA}, we can directly see why FA offers stronger defenses. 
A hypothesis here is that with the same $k$, the accuracies of base classifiers in Finite Aggregation will not change much from those in DPA,
since they have access to the same amount of training data on average and the subsets are constructed in a similar fashion. 
We have this verified empirically in Section \ref{sec:eval_certified_frac}.

With this hypothesis, we can focus on the left hand side of Theorem \ref{Th:FA} and Theorem \ref{Th:DPA} to compare the robustness of ours with DPA. By definition $1 + \text{FA}(D, x)_{c|j} - \text{FA}(D, x)_{c'|j} \leq 2$ holds for any $j\in[kd]$,
which means we always have $\frac{1}{k} \cdot \Delta_{D, x}^{\overline{d_\text{sym} (D, D')}} \leq {\frac{2}{k} \cdot d_\text{sym} (D, D')}$; thus Finite Aggregation offers better certificates than DPA. 
The intuition behind this is that when we let a sample be utilized by more than one base classifier, we can use the correlation among them to better characterize the capability of the adversary. We will later elaborate more about this insight in Section \ref{sec:IA}. 

\subsection{Practical Details}
\label{sec:PracDetail}

\textbf{The choice of $f_\text{base}$.}
The only requirement to $f_\text{base}$ is that it should be deterministic which is hardly an issue since most, if not all, classification algorithms can be made deterministic. 
Following \cite{DPA}, here we use deep neural networks for the base classifiers, where the labeled training samples are sorted in lexicographic order to remove the dependence on the order of the training set and random seeds are set explicitly. More details including model architectures can be found in Section \ref{sec:eval_setup}.

\textbf{The choice of $h_\text{split}$.}
Here we use the same hash function as \cite{DPA} for evaluation, where it simply maps each sample to $[kd]$ according to the remainder when you divide the sum of pixel values by $kd$.

\textbf{The choice of $h_\text{spread}$.} 
Here we want $h_\text{spread}$ to be a balanced hash function mapping every integer in $[kd]$ to a set of $d$ different targets in $[kd]$, where every candidate in the target space will has a preimage of the same size (i.e. $d$). The construction of $h_\text{spread}$ used in this paper is as follows:
\begin{align*}
h_\text{spread}(j) = \{ \left(j + r_t\right) \mod~kd | t \in [d]  \},
\end{align*}
where $R=\{r_0, \dots, r_{d-1}\}$ is a size-$d$ subset of $[kd]$ generated using a pseudo-random generator with a fixed random seed. One can easily verify that this is a balanced hash function when $R$ is any size-$d$ subset of $[kd]$.

\section{Analysis and Extension}
\label{sec:analysis}

\subsection{Relating to Infinite Aggregation}
\label{sec:IA}

Previously in Section \ref{sec:method}, we present Finite Aggregation as an extension of DPA \cite{DPA}.
In this section, we offer an alternative view that relates Finite Aggregation to Infinite Aggregation, which suggests the advantages of Finite Aggregation (that reduces to DPA when $d=1$) over Randomized Selection \cite{Bagging,RandomSelection}, a branch of stochastic certified defenses against data poisoning attacks. 

This insight is consistent with the observations that DPA (and therefore Finite Aggregation) typically works better than Randomized Selection empirically. For instance, on CIFAR-10, DPA \cite{DPA} can certify, with \textbf{no error rate}, more than $46\%$ of the testing samples correctly when allowing $10$ poisons and about $34\%$ when allowing $20$ poisons, while the fractions from Randomized Selection \cite{Bagging, RandomSelection} are less than $40\%$ and $25\%$ with \textbf{an error rate of $\textbf{0.1}\%$} respectively for $10$ and $20$ poisons. 
Despite the varying details (e.g. \cite{Bagging, RandomSelection} refer to a somewhat more general threat model than \cite{DPA}), all three variants are capable of dealing with supposedly the most practical threat model (i.e. poison insertions) and the gaps between DPA and the stochastic variants are fairly significant.
In addition, since the error rate from Randomized Selection is only bounded for a test set of finite size, the total error rate inevitably accumulates in deployments, where the number of test samples may increase unboundedly through time.
These are also why in Section \ref{sec:eval_certified_frac}, we benchmark Finite Aggregation against DPA to show that ours is indeed a better certified defense.

Let us take another look at the design of Finite Aggregation in Definition \ref{Def:FA}: What will happen if we let $d\to\infty$? 
Assuming the hash functions are random, 
we will have an infinite amount of subsets $S_i$, where every training sample will be spread to exactly $\frac{1}{k}$ of them independently.
This is exactly Infinite Aggregation, described as follows:

\begin{myDef}[Classification with Infinite Aggregation]
\label{Def:IA}
Given a base classifier $f_\text{base}: \mathcal{D} \times \mathcal{X} \to \mathcal{C}$, the Infinite Aggregation classifier $\text{IA}: \mathcal{D} \times \mathcal{X} \to \mathcal{C}$ is defined as follows:
\begin{align*}
    \text{IA}(D, x) = \arg \max_{c \in \mathcal{C}} 
    Pr_{S\sim \text{Bernoulli}(D, \frac{1}{k})}\left[ f_\text{base} (S, x) = c \right],
\end{align*}
where $\text{Bernoulli}(D, \frac{1}{k})$ denotes Bernoulli sampling from $D$ with sampling rate $\frac{1}{k}$, meaning that every sample in $D$ will be picked independently with a probability of $\frac{1}{k}$. 
Ties are broken by returning the smaller class index in $\arg \max$.
\end{myDef}

For simplicity, we use $\text{IA}(D, x)_{c}$ to denote the expected votes count $Pr_{S\sim \text{Bernoulli}(D, \frac{1}{k})}\left[ f_\text{base} (S, x) = c \right]$.
We use $\text{IA}(D, x)_{c|(x_L)}$ to denote the expected votes count given that sample $x_L\in D$ is utilized, which can be expressed formally as
$Pr_{S\sim \text{Bernoulli}(D \setminus \{x_L\}, \frac{1}{k})}\left[ f_\text{base} (S \cup \{x_L\}, x) = c \right]$.

Infinite Aggregation in fact is the same classifier as  Binomial selection from \cite{RandomSelection}. 
However, when we adapt the certificate from Finite Aggregation, it is quite different from theirs:

\begin{myTh}[Infinite Aggregation against Data Poisoning]
\label{Th:IA}
Given a training set $D$ and an input $x$, let $c = \text{IA}(D, x)$, for any training set $D'$, it is guaranteed that $\text{IA}(D', x) = \text{IA}(D, x)$ when for any $\delta > 0$,
\begin{align*}
    \frac{1}{k} \cdot \overline{\Delta}_{D, x}^{\overline{d_\text{sym} (D, D')}}  \leq \text{IA}(D, x)_c - \text{IA}(D, x)_{c'} - \mathbb{1}\left[c'<c\right]\cdot \delta
\end{align*}
holds for all $c' \neq c$, where $\overline{\Delta}_{D, x}$ is a multiset defined as 
\begin{align*}
\{ 1 + \text{IA}(D, x)_{c|x_L} - \text{IA}(D, x)_{c'|x_L} \}_{x_L \in D}
\\ +
\{ 1 + \text{IA}(D, x)_{c} - \text{IA}(D, x)_{c'} \} \times \infty
\end{align*}
and $\overline{\Delta}_{D, x}^{\overline{d_\text{sym} (D, D')}}$ denotes the sum of the largest $d_\text{sym} (D, D')$ elements in the multiset $\overline{\Delta}_{D, x}$. Here $+$ denotes the sum of two multisets and $\{ 1 + \text{IA}(D, x)_{c} - \text{IA}(D, x)_{c'} \} \times \infty$ denotes the multiplication of a multiset and a scalar, which is in this case a multiset containing an infinite amount of a single value, i.e. $1 + \text{IA}(D, x)_{c} - \text{IA}(D, x)_{c'}$.
\end{myTh}

The detailed proof is presented in Appendix \ref{Pf:IA}.
Note that the purpose for Theorem \ref{Th:IA} is not to propose another defense but to connect and unify different aggregation-based defenses to date.
For  \cite{Bagging,RandomSelection}, their certificates involve only the margin (e.g. $\text{IA}(D, x)_c - \text{IA}(D, x)_{c'}$) but nothing else that depends on the behavior of the base classifiers, while ours takes the advantage of fine-grained statistics through $\overline{\Delta}_{D, x}$, allowing a closer estimation of the adversary's capability.

Besides, since the number of terms in Definition \ref{Def:IA} is exponential to the training set size, Infinite Aggregation is impractical to compute exactly that an approximation of some sort will be unavoidable. \cite{Bagging,RandomSelection} do their approximations by numerically estimating the margin with Monte-Carlo methods, which introduces a probability of estimation errors that inevitably accumulates with the number of testing samples. This is not very efficient and as mentioned previously can be an issue for online services where the number of testing samples is potentially unbounded.

Finite Aggregation, however, approximates the entire scheme of Infinite Aggregation in a deterministic fashion, enabling utilizing fine-grained statistics with no error rate. 
This is not only a theoretical advantage of Finite Aggregation but may also partially explain why DPA (i.e. Finite Aggregation with $d=1$) typically works better than Randomized Selection.

\begin{figure}[!tbp]
    \centering
    \subfigure[MNIST ($k=1200$)]{
        \includegraphics[width=0.47\linewidth]{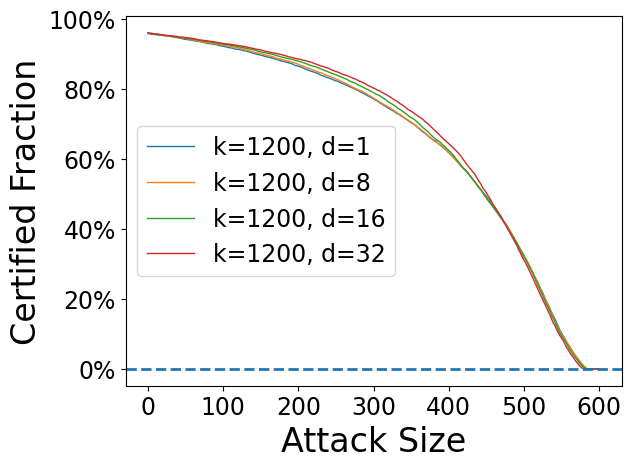}
        \hfill \label{fig:mnist}
        \includegraphics[width=0.47\linewidth]{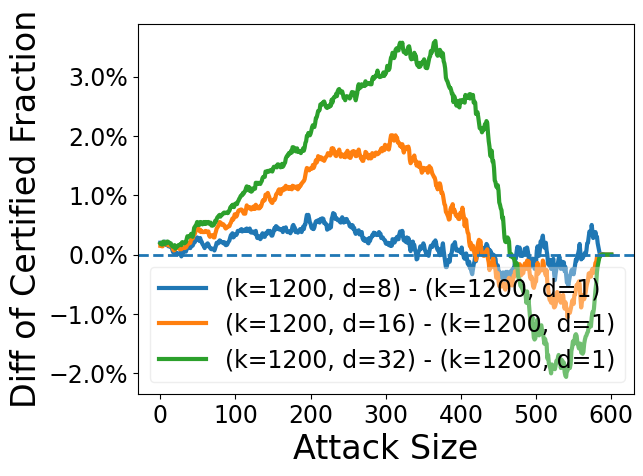}
    }
    \\
    \subfigure[CIFAR-10 ($k=50$)]{
        \includegraphics[width=0.47\linewidth]{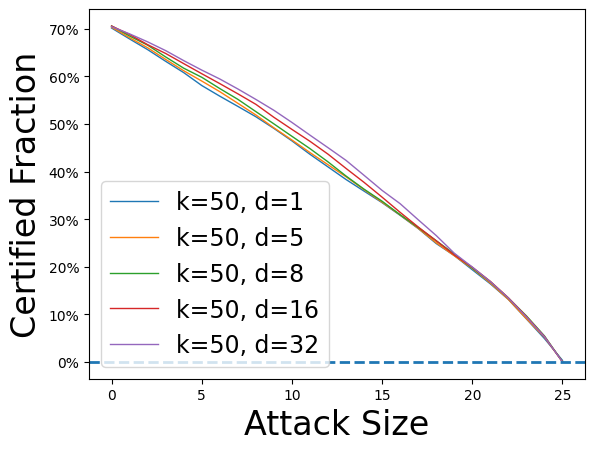}
        \hfill\label{fig:cifar_k50}
        \includegraphics[width=0.47\linewidth]{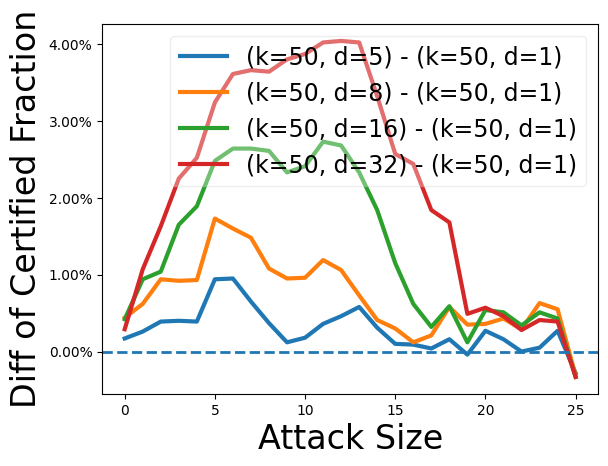}
    }
    \\
    \subfigure[CIFAR-10 ($k=250$) ]{
        \includegraphics[width=0.47\linewidth]{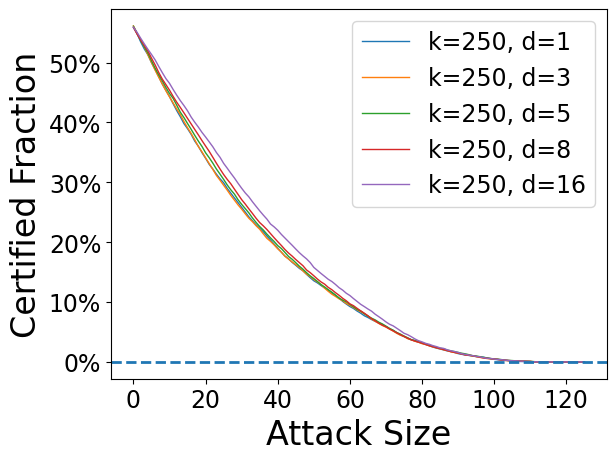}
        \hfill \label{fig:cifar_k100}
        \includegraphics[width=0.47\linewidth]{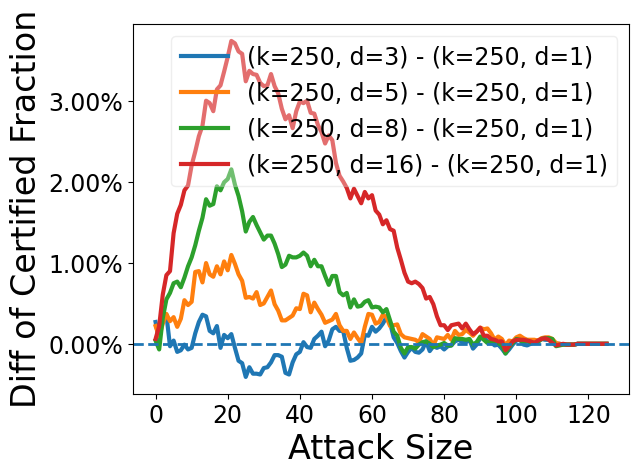}
    }
    \\
    \subfigure[GTSRB ($k=50$)]{
        \includegraphics[width=0.47\linewidth]{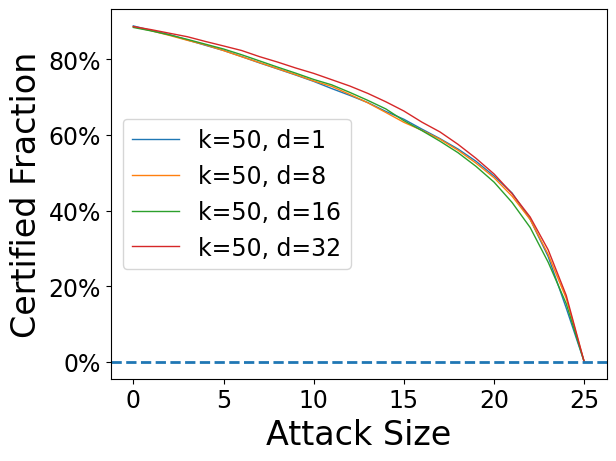}
        \hfill \label{fig:gtsrb_k50}
        \includegraphics[width=0.47\linewidth]{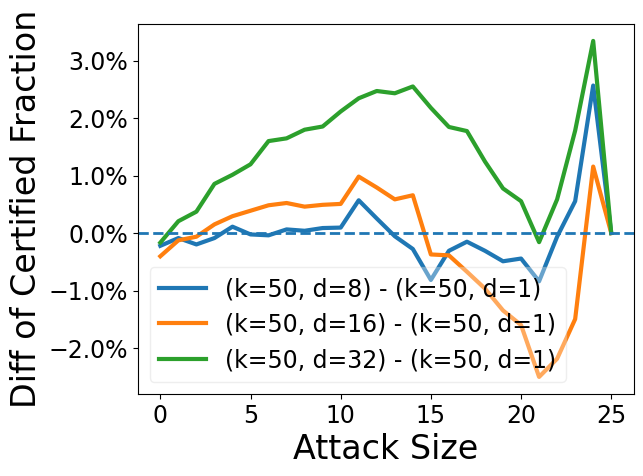}
    }
    \\
    \subfigure[GTSRB ($k=100$)]{
        \includegraphics[width=0.47\linewidth]{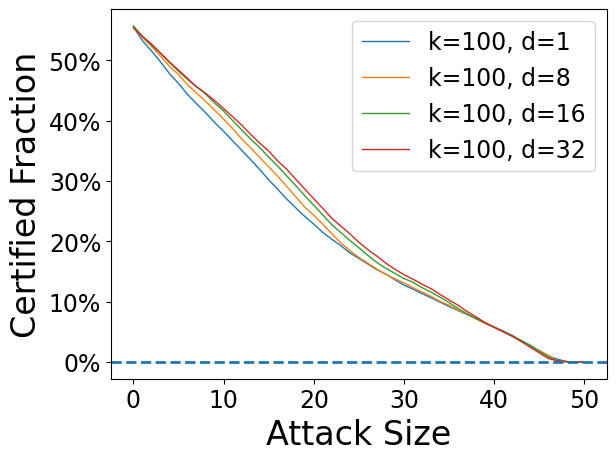}
        \hfill \label{fig:gtsrb_k100}
        \includegraphics[width=0.47\linewidth]{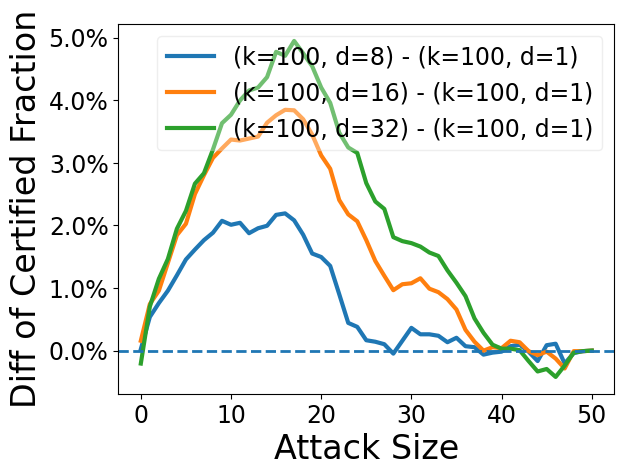}
    }
    
   \caption{ The curves of certified fraction on different datasets. \textbf{Left}: Certified fraction of Finite Aggregation with different hyperparameters. \textbf{Right}: The improvements of certified fraction from DPA (i.e. Finite Aggregation with $d=1$) to Finite Aggregation with the same $k$ and various $d$. Note that there is no direct correspondence between base classifiers with different $k$ and $d$, resulting in artifacts in the visualizations: See Figure \ref{fig:FAcerts_vs_DPAcerts_same} for the improvements using the same set of base classifiers.
   }
   \label{fig:certifed_frac}
\end{figure}

\begin{table*}[!tb]  
\renewcommand{\arraystretch}{1.2} 
\newcommand{\tablewidth}{8} 
\caption{Certified fraction of Finite Aggregation with various hyperparameters with respect to different attack sizes $d_\text{sim}$. The improvements compared to the DPA baseline are highlighted in \textcolor{blue}{blue} if they are positive and \textcolor{red}{red} otherwise. Note that there is no direct correspondence between base classifiers with different $k$ and $d$, resulting in artifacts in the visualizations: See Figure \ref{fig:FAcerts_vs_DPAcerts_same} for the improvements using the same set of base classifiers.} 
\label{tab:certified_frac}
\vskip 0.1in
\centering
\resizebox{1\linewidth}{!}{ 
\begin{tabular}{|c||c||c||l|l|l|l|l|l}
\hline
\multicolumn{1}{|c||}{\bfseries dataset} & \bfseries k & \bfseries d & \multicolumn{ 5 }{c|}{\bfseries certified fraction} \\ \hline\hline
\multicolumn{1}{|c||}{\multirow{5}{*}{MNIST}} & \multicolumn{1}{|c||}{\multirow{5}{*}{1200}} &  & \multicolumn{1}{c|}{$d_\text{sim} \leq 50$} & \multicolumn{1}{c|}{$d_\text{sim} \leq 100$} & \multicolumn{1}{c|}{$d_\text{sim} \leq 200$} & \multicolumn{1}{c|}{$d_\text{sim} \leq 300$} & \multicolumn{1}{c|}{$d_\text{sim} \leq 400$} \\ \cline{3-\tablewidth}
\multicolumn{1}{|c||}{}&   & 1 (DPA) & $94.12\%$ & $92.11\%$ & $86.45\%$ & $77.12\%$ & $61.78\%$\\ \cline{3-\tablewidth}
\multicolumn{1}{|c||}{}&   & 8 & $94.38\%\color{blue}{(+0.26\%)}$ & $92.45\%\color{blue}{(+0.34\%)}$ & $86.97\%\color{blue}{(+0.52\%)}$ & $77.31\%\color{blue}{(+0.19\%)}$ & $61.81\%\color{blue}{(+0.03\%)}$ \\ \cline{3-\tablewidth}
\multicolumn{1}{|c||}{}&   & 16 & $94.54\%\color{blue}{(+0.42\%)}$ & $92.75\%\color{blue}{(+0.64\%)}$ & $87.89\%\color{blue}{(+1.44\%)}$ & $78.91\%\color{blue}{(+1.79\%)}$ & $62.42\%\color{blue}{(+0.64\%)}$ \\ \cline{3-\tablewidth}
\multicolumn{1}{|c||}{}&  & 32 & $94.63\%\color{blue}{(+0.51\%)}$ & $92.97\%\color{blue}{(+0.86\%)}$ & $88.49\%\color{blue}{(+2.04\%)}$ & $80.17\%\color{blue}{(+3.05\%)}$ & $64.34\%\color{blue}{(+2.56\%)}$ \\ 
\hline
\hline
\multicolumn{1}{|c||}{\multirow{12}{*}{CIFAR-10}} & \multicolumn{1}{|c||}{\multirow{6}{*}{50}} & & \multicolumn{1}{c|}{$d_\text{sim} \leq 3$} & \multicolumn{1}{c|}{$d_\text{sim} \leq 5$} & \multicolumn{1}{c|}{$d_\text{sim} \leq 10$} & \multicolumn{1}{c|}{$d_\text{sim} \leq 15$} & \multicolumn{1}{c|}{$d_\text{sim} \leq 20$} \\ \cline{3-\tablewidth}
\multicolumn{1}{|c||}{}&  & 1 (DPA) & $63.15\%$ & $58.07\%$ & $46.44\%$ & $33.46\%$ & $19.36\%$\\ \cline{3-\tablewidth}
\multicolumn{1}{|c||}{}&  & 5 & $63.55\%\color{blue}{(+0.40\%)}$ & $59.01\%\color{blue}{(+0.94\%)}$ & $46.62\%\color{blue}{(+0.18\%)}$& $33.56\%\color{blue}{(+0.10\%)}$ & $19.63\%\color{blue}{(+0.27\%)}$ \\ \cline{3-\tablewidth}
\multicolumn{1}{|c||}{}&  &  8 & $64.07\%\color{blue}{(+0.92\%)}$ & $59.80\%\color{blue}{(+1.73\%)}$ & $47.40\%\color{blue}{(+0.96\%)}$& $33.76\%\color{blue}{(+0.30\%)}$ & $19.72\%\color{blue}{(+0.36\%)}$ \\ \cline{3-\tablewidth}
\multicolumn{1}{|c||}{}&  & 16 & $64.80\%\color{blue}{(+1.65\%)}$ & $60.55\%\color{blue}{(+2.48\%)}$ & $48.85\%\color{blue}{(+2.41\%)}$& $34.61\%\color{blue}{(+1.15\%)}$ & $19.90\%\color{blue}{(+0.54\%)}$ \\ \cline{3-\tablewidth}
\multicolumn{1}{|c||}{}&  & 32 & $65.40\%\color{blue}{(+2.25\%)}$ & $61.31\%\color{blue}{(+3.24\%)}$ & $50.31\%\color{blue}{(+3.87\%)}$ & $36.03\%\color{blue}{(+2.57\%)}$ & $19.93\%\color{blue}{(+0.57\%)}$ \\ 
\cline{2-\tablewidth}
\noalign{\vskip\doublerulesep
         \vskip-\arrayrulewidth}
\cline{2-\tablewidth}
 & \multicolumn{1}{|c||}{\multirow{6}{*}{250}} & & \multicolumn{1}{c|}{$d_\text{sim} \leq 10$} & \multicolumn{1}{c|}{$d_\text{sim} \leq 20$} & \multicolumn{1}{c|}{$d_\text{sim} \leq 30$} & \multicolumn{1}{c|}{$d_\text{sim} \leq 40$} & \multicolumn{1}{c|}{$d_\text{sim} \leq 50$} \\ \cline{3-\tablewidth}
\multicolumn{1}{|c||}{}&  & 1 (DPA) & $44.31\%$ & $34.01\%$ & $25.81\%$ & $18.99\%$ & $13.55\%$\\ \cline{3-\tablewidth}
\multicolumn{1}{|c||}{}&  & 3 & $44.26\%\color{red}{(-0.05\%)}$ & $34.08\%\color{blue}{(+0.07\%)}$ & $25.51\%\color{red}{(-0.30\%)}$& $18.89\%\color{red}{(-0.10\%)}$ & $13.76\%\color{blue}{(+0.21\%)}$ \\ \cline{3-\tablewidth}
\multicolumn{1}{|c||}{}&  & 5 & $44.83\%\color{blue}{(+0.52\%)}$ & $34.92\%\color{blue}{(+0.91\%)}$ & $26.31\%\color{blue}{(+0.50\%)}$& $19.42\%\color{blue}{(+0.43\%)}$ & $13.92\%\color{blue}{(+0.37\%)}$ \\ \cline{3-\tablewidth}
\multicolumn{1}{|c||}{}&  & 8 & $45.38\%\color{blue}{(+1.07\%)}$ & $36.05\%\color{blue}{(+2.04\%)}$ & $27.10\%\color{blue}{(+1.29\%)}$& $20.08\%\color{blue}{(+1.09\%)}$ & $14.39\%\color{blue}{(+0.84\%)}$ \\ \cline{3-\tablewidth}
\multicolumn{1}{|c||}{}&  & 16 & $46.52\%\color{blue}{(+2.21\%)}$ & $37.56\%\color{blue}{(+3.55\%)}$ & $29.00\%\color{blue}{(+3.19\%)}$ & $22.00\%\color{blue}{(+3.01\%)}$ & $15.79\%\color{blue}{(+2.24\%)}$ \\ 
\hline\hline
\multicolumn{1}{|c||}{\multirow{10}{*}{GTSRB}} & \multicolumn{1}{|c||}{\multirow{5}{*}{50}} & & \multicolumn{1}{c|}{$d_\text{sim} \leq 3$} & \multicolumn{1}{c|}{$d_\text{sim} \leq 5$} & \multicolumn{1}{c|}{$d_\text{sim} \leq 10$} & \multicolumn{1}{c|}{$d_\text{sim} \leq 15$} & \multicolumn{1}{c|}{$d_\text{sim} \leq 24$} \\ \cline{3-\tablewidth}
\multicolumn{1}{|c||}{}&  & 1 (DPA) & $85.09\%$ & $82.32\%$ & $74.15\%$ & $64.14\%$ & $14.27\%$\\ \cline{3-\tablewidth}
\multicolumn{1}{|c||}{}&  & 8 & $85.00\%\color{red}{(-0.09\%)}$ & $82.30\%\color{red}{(-0.02\%)}$ & $74.24\%\color{blue}{(+0.09\%)}$& $63.33\%\color{red}{(-0.81\%)}$ & $16.83\%\color{blue}{(+2.56\%)}$ \\ \cline{3-\tablewidth}
\multicolumn{1}{|c||}{}&  & 16 & $85.25\%\color{blue}{(+0.16\%)}$ & $82.71\%\color{blue}{(+0.39\%)}$ & $74.66\%\color{blue}{(+0.51\%)}$& $63.77\%\color{red}{(-0.37\%)}$ & $15.42\%\color{blue}{(+1.15\%)}$ \\ \cline{3-\tablewidth}
\multicolumn{1}{|c||}{}&  & 32 & $85.95\%\color{blue}{(+0.86\%)}$ & $83.52\%\color{blue}{(+1.20\%)}$ & $76.26\%\color{blue}{(+2.11\%)}$ & $66.32\%\color{blue}{(+2.18\%)}$ & $17.61\%\color{blue}{(+3.34\%)}$ \\
\cline{2-\tablewidth}
\noalign{\vskip\doublerulesep
         \vskip-\arrayrulewidth}
\cline{2-\tablewidth}
 & \multicolumn{1}{|c||}{\multirow{5}{*}{100}} & & \multicolumn{1}{c|}{$d_\text{sim} \leq 5$} & \multicolumn{1}{c|}{$d_\text{sim} \leq 10$} & \multicolumn{1}{c|}{$d_\text{sim} \leq 15$} & \multicolumn{1}{c|}{$d_\text{sim} \leq 20$} & \multicolumn{1}{c|}{$d_\text{sim} \leq 25$} \\ \cline{3-\tablewidth}
\multicolumn{1}{|c||}{}&  & 1 (DPA) & $46.16\%$ & $38.24\%$ & $30.19\%$ & $22.84\%$ & $17.16\%$\\ \cline{3-\tablewidth}
\multicolumn{1}{|c||}{}&  & 8 & $47.62\%\color{blue}{(+1.46\%)}$ & $40.25\%\color{blue}{(+2.01\%)}$ & $32.36\%\color{blue}{(+2.17\%)}$& $24.34\%\color{blue}{(+1.50\%)}$ & $17.32\%\color{blue}{(+0.16\%)}$ \\ \cline{3-\tablewidth}
\multicolumn{1}{|c||}{}&  & 16 & $48.19\%\color{blue}{(+2.03\%)}$ & $41.62\%\color{blue}{(+3.38\%)}$ & $33.95\%\color{blue}{(+3.76\%)}$& $25.96\%\color{blue}{(+3.12\%)}$ & $18.92\%\color{blue}{(+1.76\%)}$ \\ \cline{3-\tablewidth}
\multicolumn{1}{|c||}{}&  & 32 & $48.39\%\color{blue}{(+2.23\%)}$ & $42.01\%\color{blue}{(+3.77\%)}$ & $34.96\%\color{blue}{(+4.77\%)}$ & $27.05\%\color{blue}{(+4.21\%)}$ & $19.93\%\color{blue}{(+2.77\%)}$ \\
\hline
\end{tabular}
}
\vskip -0.1in
\end{table*}

\begin{table}[!tb]  
\renewcommand{\arraystretch}{1.15} 
\newcommand{\tablewidth}{6} 
\caption{Some statistics of Finite Aggregation, where $\text{acc}_\text{clean}$ denotes the test accuracy with a clean training set; $\text{acc}_\text{base}$ denotes the average accuracy of base classifiers; $Pr[ r \uparrow ]$ denotes the fraction of samples in the testing set that obtain a larger certified radius when using our certificates from Theorem \ref{Th:FA} instead of the ones from DPA; and $\Delta r$ denotes the average increase of the certified radius among those getting a larger radius.} 
\label{tab:stats}
\vskip 0.1in
\centering
\resizebox{1\linewidth}{!}{ 
\begin{tabular}{c|c|c|cccc}
\hline
 dataset &  k &  d & $\text{acc}_\text{clean}$ & $\text{acc}_\text{base}$ & $Pr[ r \uparrow ]$ & $\Delta r$ \\ \hline
\multirow{4}{*}{MNIST} & \multirow{4}{*}{1200} & 1 (DPA) & $95.75\%$ &$76.92\%$ & $0\%$ & $0$\\
&  & 8 & $95.95\%$ &$76.92\%$ & $15.72\%$ & $6.96$ \\
&  & 16 & $95.90\%$ & $76.83\%$ &$35.24\%$ & $12.69$ \\
&  & 32 & $95.94\%$ &$76.54\%$ & $58.01\%$ & $17.91$\\
\hline
\multirow{10}{*}{CIFAR-10} & \multirow{5}{*}{50} & 1 (DPA) & $70.15\%$ &$56.15\%$ & $0\%$ & $0$\\
&  & 5 & $70.32\%$ &$56.14\%$ & $1.49\%$ & $1.00$ \\
&  & 8 & $70.59\%$ & $56.33\%$ &$6.67\%$ & $1.00$ \\
&  & 16 & $70.57\%$ &$56.32\%$ & $22.68\%$ & $1.02$\\
&  & 32 & $70.44\%$ &$56.27\%$ & $37.92\%$ & $1.18$\\
\cline{2-7}
 & \multirow{5}{*}{250} & 1 (DPA) & $55.84\%$ &$35.21\%$ & $0\%$ & $0$\\
&  & 3 & $56.11\%$ &$35.15\%$ & $0.33\%$ & $1.03$ \\
&  & 5 & $56.06\%$ & $35.24\%$ &$15.32\%$ & $1.30$ \\
&  & 8 & $55.88\%$ &$35.18\%$ & $36.07\%$ & $1.75$\\
&  & 16 & $55.90\%$ &$35.24\%$ & $50.40\%$ & $3.21$\\
\hline
\multirow{8}{*}{GTSRB} & \multirow{4}{*}{50} & 1 (DPA) & $88.80\%$ &$74.47\%$ & $0\%$ & $0$\\
&  & 8 & $88.58\%$ & $73.71\%$ &$6.50\%$ & $1.00$ \\
&  & 16 & $88.40\%$ &$73.09\%$ & $18.58\%$ & $1.03$\\
&  & 32 & $88.64\%$ &$73.92\%$ & $29.66\%$ & $1.20$\\
\cline{2-7}
 & \multirow{4}{*}{100} & 1 (DPA) & $55.56\%$ &$34.71\%$ & $0\%$ & $0$\\
&  & 8 & $55.58\%$ &$34.55\%$ & $29.39\%$ & $1.16$\\
&  & 16 & $55.72\%$ &$34.58\%$ & $41.50\%$ & $1.74$\\
&  & 32 & $55.35\%$ &$34.20\%$ & $46.71\%$ & $2.41$\\
\hline
\end{tabular}
}
\vskip -0.1in
\end{table}

\begin{table}[!tb]  
\renewcommand{\arraystretch}{1.15} 
\newcommand{\tablewidth}{6} 
\caption{Certified fraction ($\text{frac}_\text{certified}$) and certified accuracy ($\text{acc}_\text{certified}$) of Finite Aggregation corresponding to an attack size of 1. 
Their differences are highlighted in \textcolor{blue}{blue}.
}
\label{tab:cert_acc}
\vskip 0.1in
\centering
\resizebox{1\linewidth}{!}{ 
\begin{tabular}{c|c|c|ccc}
\hline
 dataset &  k &  d & $\text{acc}_\text{clean}$ & $\text{frac}_\text{certified}$ & $\text{acc}_\text{certified}$ \\ \hline
\multirow{4}{*}{MNIST} & \multirow{4}{*}{1200} & 1 (DPA) & $95.75\%$ &$95.71\%$ & $95.71\%\color{blue}{(+0\%)}$ \\
&  & 8 & $95.95\%$ &$95.91\%$ & $95.91\% \color{blue}{(+0\%)}$ \\
&  & 16 & $95.90\%$ & $95.86\%$ &$95.86\%\color{blue}{(+0\%)}$  \\
&  & 32 & $95.94\%$ &$95.91\%$ & $95.91\%\color{blue}{(+0\%)}$ \\
\hline
\multirow{10}{*}{CIFAR-10} & \multirow{5}{*}{50} & 1 (DPA) & $70.15\%$ &$67.85\%$ & $68.79\%\color{blue}{(+0.94\%)}$ \\
&  & 5 & $70.32\%$ &$68.11\%$ & $69.07\%\color{blue}{(+0.96\%)}$ \\
&  & 8 & $70.59\%$ & $68.47\%$ &$69.22\%\color{blue}{(+0.75\%)}$  \\
&  & 16 & $70.57\%$ &$68.79\%$ & $69.38\%\color{blue}{(+0.59\%)}$ \\
&  & 32 & $70.44\%$ &$68.92\%$ & $69.34\%\color{blue}{(+0.42\%)}$\\
\cline{2-6}
 & \multirow{5}{*}{250} & 1 (DPA) & $55.84\%$ &$54.82\%$ & $55.19\%\color{blue}{(+0.37\%)}$ \\
&  & 3 & $56.11\%$ &$54.86\%$ & $55.37\%\color{blue}{(+0.51\%)}$ \\
&  & 5 & $56.06\%$ & $54.91\%$ &$55.34\%\color{blue}{(+0.43\%)}$  \\
&  & 8 & $55.88\%$ &$54.75\%$ & $55.13\%\color{blue}{(+0.38\%)}$ \\
&  & 16 & $55.90\%$ &$55.04\%$ & $55.35\%\color{blue}{(+0.29\%)}$ \\
\hline
\multirow{8}{*}{GTSRB} & \multirow{4}{*}{50} & 1 (DPA) & $88.80\%$ &$87.61\%$ & $87.99\%\color{blue}{(+0.38\%)}$ \\
&  & 8 & $88.58\%$ & $87.52\%$ &$87.88\%\color{blue}{(+0.36\%)}$ \\
&  & 16 & $88.40\%$ &$87.48\%$ & $87.74\%\color{blue}{(+0.26\%)}$ \\
&  & 32 & $88.64\%$ &$87.81\%$ & $87.97\%\color{blue}{(+0.16\%)}$ \\
\cline{2-6}
 & \multirow{4}{*}{100} & 1 (DPA) & $55.56\%$ &$53.26\%$ & $53.96\%\color{blue}{(+0.70\%)}$\\
&  & 8 & $55.58\%$ &$53.80\%$ & $54.39\%\color{blue}{(+0.59\%)}$ \\
&  & 16 & $55.72\%$ &$54.01\%$ & $54.49\%\color{blue}{(+0.48\%)}$ \\
&  & 32 & $55.35\%$ &$53.97\%$ & $54.25\%\color{blue}{(+0.28\%)}$ \\
\hline
\end{tabular}
}
\vskip -0.1in
\end{table}

\begin{figure}[!tb]
    \centering
    \subfigure[MNIST (k=1200)]{
        \includegraphics[width=0.47\linewidth]{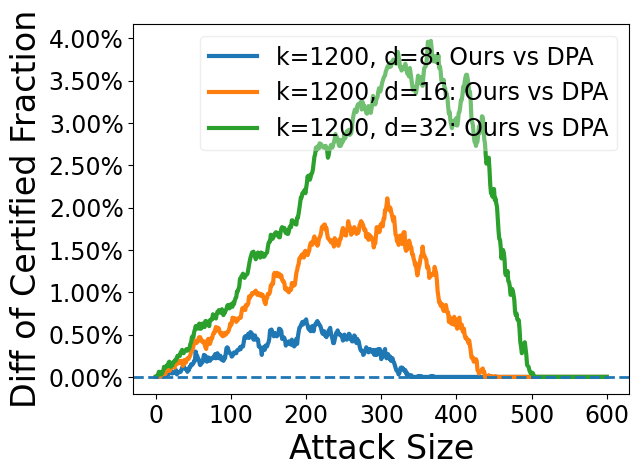}
    }
    \hfill
    \subfigure[CIFAR-10 (k=50)]{
        \includegraphics[width=0.47\linewidth]{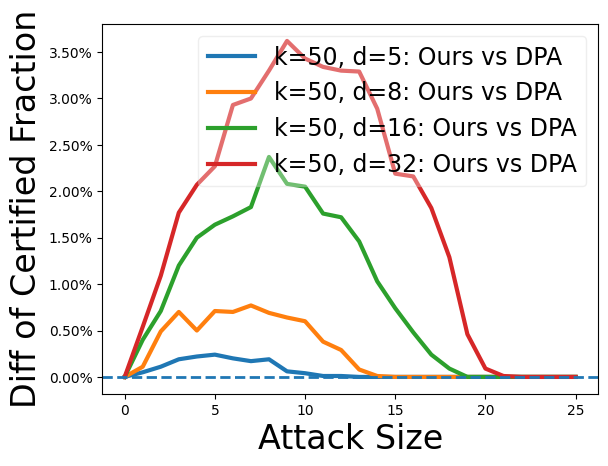}
    }
    
     \subfigure[CIFAR-10 (k=250)]{
         \includegraphics[width=0.47\linewidth]{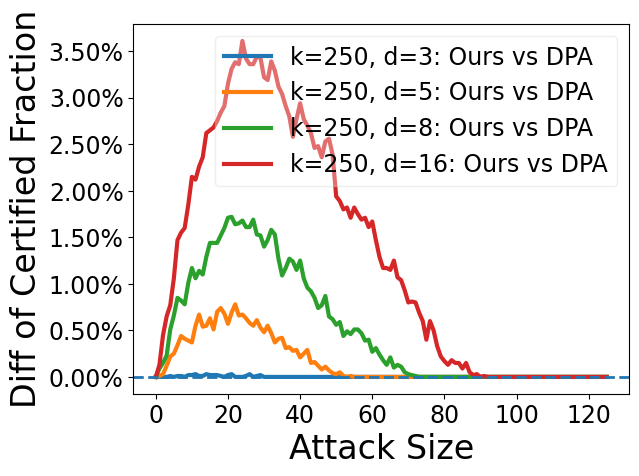}
     }
     \hfill
     \subfigure[GTSRB (k=50)]{
         \includegraphics[width=0.47\linewidth]{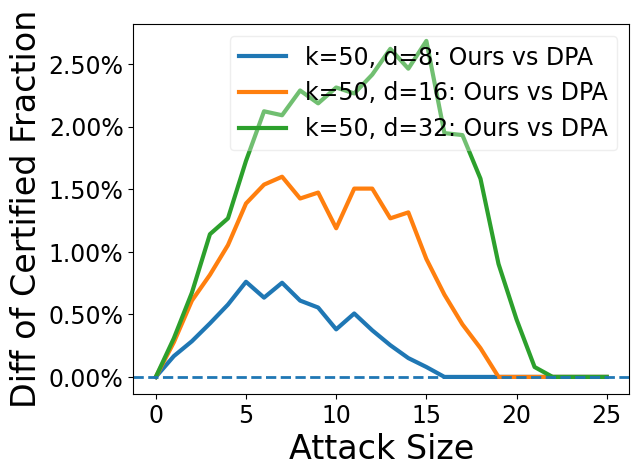}
     }
     
    \subfigure[GTSRB (k=100)]{
        \includegraphics[width=0.47\linewidth]{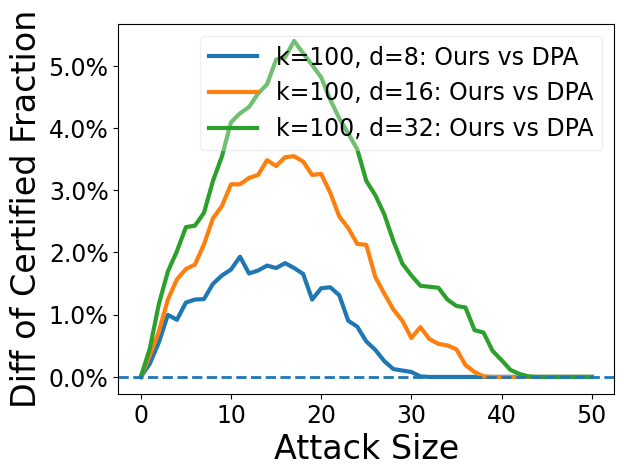}
    }
    \hfill

   \caption{The improvements of certified fraction when applying our certificates (Theorem \ref{Th:FA}) instead of DPA certificates to the same set of $k\cdot d$ base classifiers.
   }
   
   \label{fig:FAcerts_vs_DPAcerts_same}
\end{figure}

\section{Evaluation}
\label{sec:eval}

\subsection{Evaluation Setup}
\label{sec:eval_setup}

We follow exactly the setup of \cite{DPA} in the experiments and use the same hyperparameters as theirs. As mentioned in Section \ref{sec:IA}, we compare our method with DPA since it is empirically the state of the art in certified defenses against general poisoning attacks.

\textbf{Datasets.} We evaluate our method on MNIST \cite{MNIST}, CIFAR-10 \cite{cifar} and GTSRB \cite{GTSRB} datasets, which are respectively 10-way classification of handwritten digits, 10-way object classification and 43-way classification of traffic signs.

\textbf{Training hyperparameters.} We use the Network-In-Network \cite{NIN} architecture, trained with the hyperparameters from \cite{RotNet}.
On MNIST and GTSRB, we also exclude horizontal flips in data augmentations as in \cite{DPA}.

\subsection{Certified Predictions}
\label{sec:eval_certified_frac}

We use {\it certified fraction} as our performance metric, which refers to the fraction of samples in the testing set that are not only correctly classified but also certified to be robust given a certain attack size. 
This is the same metric as `certified accuracy' in \cite{DPA} but we choose to refer to it differently.
Our motivation is discussed in Section \ref{sec:cert_acc}.

\textbf{Improved certified robustness.} We report the certified fractions of Finite Aggregation in Table \ref{tab:certified_frac} and Figure \ref{fig:certifed_frac}. Overall, it is evident that Finite Aggregation can offer strong certified defenses than DPA, where the improvements of certified fractions can be up to $3\%$ or $4\%$ compared to DPA using the same hyperparameters.

Now we examine the effectiveness of our certificates through the scope of certified radius. Given a test sample $x$, the certified radius is simply the maximum attack size $d_\text{sym}$ allowed while we can still certify the correct prediction on $x$. The certified radius is considered to be $-1$ if the prediction on a test sample does not match the true label.

In Table \ref{tab:stats}, we include two statistics relating to certified radii: $Pr[ r\uparrow ]$, which denotes the fraction of samples in the testing set that obtain a larger certified radius when using our certificates from Theorem \ref{Th:FA} instead of the ones from DPA (i.e. replacing $\Delta_{D, x}^{\overline{d_\text{sym} (D, D')}}$ in Theorem \ref{Th:FA} with $2\cdot d_\text{sym} (D, D')$), 
and $\Delta r$, which denotes the average increase of the certified radius among those getting a larger radius.
See Figure \ref{fig:FAcerts_vs_DPAcerts_same} for the corresponding certified fraction.
The values of $Pr[ r \uparrow ]$ and $\Delta r$ in Table \ref{tab:stats} are strong supports to the effectiveness of our certificates. 

We also report in Table \ref{tab:stats} respectively the accuracy of Finite Aggregation $\text{acc}_\text{clean}$ and the average accuracy of base classifiers $\text{acc}_\text{base}$, which supports our hypothesis in Section \ref{sec:robustness} that the accuracies of base classifiers will not change much.

\subsection{Hyperparameters for Finite Aggregation}
In this section, we discuss how the hyperparameters $k$ and $d$ affect the behaviors of Finite Aggregation in practice.

\textbf{The effect of $k$: accuracy vs. robustness.}
Similar to DPA, $k$ corresponds to a trade-off between accuracy and robustness. Since every base classifier will on average have access to $1/k$ of the training set, using a larger $k$ will reduce the accuracies of base classifiers and therefore restrict the accuracy of the aggregation. 
However, in Theorem \ref{Th:FA}, the sensitivity of the normalized margin to every poisoned sample is also proportional to $1/k$, suggesting that a larger $k$ can reduce the sensitivity and favor robustness.
This is confirmed in Table \ref{tab:certified_frac} and Table \ref{tab:stats}, where a larger $k$ typically corresponds to worse accuracy $\text{acc}_\text{clean}$ but may offer a higher certified fraction when the attack sizes are relatively large.

\textbf{The effect of $d$: efficiency vs. robustness.}
The spreading degree $d$ controls how many base classifiers will have access to the same training sample.
From Table \ref{tab:certified_frac}, we can see that increasing $d$ tends to improve the robustness of Finite Aggregation, indicated by the increase of certified fractions. However, unlike $k$, increasing $d$ will not degrade the accuracy of Finite Aggregation, as observed in Table \ref{tab:stats}.
The major cost of using a larger $d$ is the increase of computation, because the number of base classifiers is proportional to $d$ while the average number of training samples for every base classifier does not depend on $d$.
Thus $d$ actually allows us to obtain stronger robustness at a cost of extra computation.

Another effect of $d$ is that using a larger $d$ tends to avoid ties in the aggregation, as indicated by the term $\frac{\mathbb{1}\left[c'<c\right]}{k d}$ in Theorem \ref{Th:FA}.
This can be quite beneficial depending on tasks. 
For instance, in Figure \ref{fig:gtsrb_k50} and in Table \ref{tab:certified_frac}, one can notice extra improvements of certified fraction from DPA for an attack size of $24$ on GTSRB with $k=50$, which exactly attributes to the reduction of ties when $d$ increases.

\subsection{Pointwise Robustness vs Distributional Robustness}
\label{sec:cert_acc}
In this section, we explore a metric corresponding to distributional robustness, namely certified accuracy, that is the lower bound on accuracy in the test set under poisoning attacks.
This is different from {\it certified fraction} in Section \ref{sec:eval_certified_frac}, which denotes the fraction of samples in the testing set that are certified to be correct under poisoning attacks:
By definition, certified fraction is smaller than certified accuracy.
For instance, when there are only two test samples, the adversarial attack budget (i.e., the number of poisons) may be enough to flip the prediction on either sample, but not enough to flip both predictions simultaneously, resulting in a certified fraction of $0\%$ and a certified accuracy of $50\%$.

While our method is designed for pointwise robustness (i.e. corresponding to certified fraction), it naturally offers a well-defined but computationally inefficient way to estimate certified accuracy (Appendix \ref{appendix:FA_cert_acc}). 
Taking advantage of this, we directly compare certified accuracy and certified fraction under an attack strength $d_\text{sym} = 1$ in Table \ref{tab:cert_acc} to estimate their gaps.
The differences in Table \ref{tab:cert_acc} corroborate the intuition that different poisons may be needed for different targets.
Notably, to our best knowledge, this is one of the first {\it direct} empirical comparisons between pointwise and distributional certified robustness resulting from a {\it single} method. A concurrent work \cite{CollectiveRobust} highlights the distributional robustness of aggregation-based defenses.

\section{Conclusion}
\label{sec:conclusion}

In this work, we propose \textbf{Finite Aggregation}, a novel provable defense against general data poisoning extended from DPA to further improve robustness.
Compared to DPA, our method effectively boosts certified fractions by up to $3.05\%$, $3.87\%$ and $4.77\%$ on MNIST, CIFAR-10, and GTSRB, respectively, 
achieving a new \textbf{state of the art} in pointwise certified robustness against general data poisoning.
We also provide an alternative view to aggregation-based defenses against poisoning attacks that bridges the gap between the deterministic and the stochastic variants, unifying the designs of aggregation-based defenses to date. 

\section*{Acknowledgements}
This project was supported in part by NSF CAREER AWARD 1942230, a grant from NIST 60NANB20D134, HR001119S0026 (GARD), ONR YIP award N00014-22-1-2271 and Army Grant No. W911NF2120076.


\bibliography{ref}
\bibliographystyle{icml2022}

\appendix

\section{Detailed Explanation of Figure \ref{fig:FA_toy}}
\label{sec:toy_example_explain}

Figure \ref{fig:FA_toy} is a toy example illustrating how our proposed Finite Aggregation improves provable robustness through a strategic reusing of every sample. 
The correct prediction is `cat' in this example.
When there is no poison, both DPA and our method predict correctly with the same distribution of predictions from base classifiers (i.e. $50\%$ cat, $16.7\%$ dog, $16.7\%$ deer, and $16.7\%$ bird).
In DPA, with one poison contributes to a base classifier that originally predicts correctly (i.e. `cat'), it may reduce the margin between `cat' and `dog' by $2/6 = 33.3\%$ and create a tie, as shown in Figure \ref{fig:FA_toy}.

However, with Finite Aggregation, even the most effective poison cannot be as effective.
In the example, every partition contributes to two consecutive base classifiers and the predictions of them happen to be different.
This means that one poison will never remove two correct predictions (i.e. `cat') from base classifiers and therefore can at most remove 1 `cat' and add 2 `dog' (or other classes) to reduce the margin by $3/12 = 25\%$, which is still too small to affect our final prediction.


\section{Proof of Theorem \ref{Th:FA}}
\label{Pf:FA}
\begin{proof}

Given a training set $D$ and an input $x$, for an arbitrary training set $D'$ such that
\begin{align*}
    \frac{1}{k} \cdot \Delta_{D, x}^{\overline{d_\text{sym} (D, D')}}  \leq \text{FA}(D, x)_c - \text{FA}(D, x)_{c'} - \frac{\mathbb{1}\left[c'<c\right]}{kd}
\end{align*}
holds for all $c' \neq c$, we want to show that $\text{FA}(D', x) = \text{FA}(D, x) = c$.

Let $D \ominus D' = (D\setminus D')\cup (D'\setminus D)$ to denote symmetric difference between $D$ and $D'$, which corresponds to the minimum set of samples to be inserted/removed if one wants change from $D$ to $D'$ (or $D'$ to $D$).

Since the base classification algorithm $f_\text{base}$ is deterministic, by definition, the prediction from a base classifier will not change if its corresponding training set $S_i$ ($S_i'$) remains unchanged from $D$ to $D'$, which is equivalent to $i \notin \bigcup_{x \in D\ominus D'} h_\text{spread}(h_\text{split(x)})$.
Let $$h_{D\ominus D'} = \bigcup_{x \in D\ominus D'} h_\text{spread}(h_\text{split(x)}).$$ 

Thus for any $c' \neq c$, we have
\begin{align*}
    &\text{FA}(D', x)_c - \text{FA}(D', x)_{c'}\\
    &=\frac{1}{kd} \sum_{i\in [kd]}\Big( \mathbb{1}\left[ 
    f_\text{base}(S'_i, x)=c\right] - \mathbb{1}\left[ 
    f_\text{base}(S'_i, x)=c'\right] \Big)\\
    &= \text{FA}(D, x)_c - \text{FA}(D, x)_{c'}  \\
    &+\frac{1}{kd} 
     \sum_{i\in h_{D\ominus D'}}\Big( \mathbb{1}\left[ 
    f_\text{base}(S'_i, x)=c\right] - \mathbb{1}\left[ 
    f_\text{base}(S'_i, x)=c'\right]  \Big)  \\
    &-\frac{1}{kd} 
     \sum_{i\in h_{D\ominus D'}} \Big( \mathbb{1}\left[ 
    f_\text{base}(S_i, x)=c\right] - \mathbb{1}\left[ 
    f_\text{base}(S_i, x)=c'\right]  \Big)\\
    &\geq \text{FA}(D, x)_c - \text{FA}(D, x)_{c'}  \\
    &+\frac{1}{kd} 
     \sum_{i\in h_{D\ominus D'}} \Big( 0 - 1  \Big)  \\
    &-\frac{1}{kd} 
    \sum_{i\in h_{D\ominus D'}} \Big( \mathbb{1}\left[ 
    f_\text{base}(S_i, x)=c\right] - \mathbb{1}\left[ 
    f_\text{base}(S_i, x)=c'\right]  \Big)\\
    & = \text{FA}(D, x)_c - \text{FA}(D, x)_{c'} - \\
    &\frac{1}{kd} 
     \sum_{i\in h_{D\ominus D'}} \Big(1 + \mathbb{1}\left[ 
    f_\text{base}(S_i, x)=c\right] - \mathbb{1}\left[ 
    f_\text{base}(S_i, x)=c'\right]  \Big)\\
\end{align*}

We can rewrite $h_{D\ominus D'}$ as follows:
\begin{align*}
h_{D\ominus D'} = \bigcup_{x \in D\ominus D'} h_\text{spread}(h_\text{split(x)}) = \bigcup_{j \in h_\text{split}(D\ominus D')} h_\text{spread}(j),
\end{align*}
where $h_\text{split}(D\ominus D') = \{ h_\text{split}(x) | x\in D\ominus D' \}$.

Since $1 + \mathbb{1}\left[ f_\text{base}(S_i, x)=c\right] - \mathbb{1}\left[ f_\text{base}(S_i, x)=c'\right] \geq 0$,
we can further bound the above formula as follows:
\begin{align*}
&\text{FA}(D', x)_c - \text{FA}(D', x)_{c'}\\
&\geq  \text{FA}(D, x)_c - \text{FA}(D, x)_{c'}  \\
&-\frac{1}{kd} 
 \sum_{j\in h_\text{split}(D \ominus D')}\Big( \sum_{i \in h_\text{spread}(j)} 1 + \mathbb{1}\left[ 
f_\text{base}(S_i, x)=c\right]  \\
& - \mathbb{1}\left[ 
f_\text{base}(S_i, x)=c'\right]  \Big)\\
&=\text{FA}(D, x)_c - \text{FA}(D, x)_{c'}  \\
&-\frac{1}{k} 
 \sum_{j\in h_\text{split}(D \ominus D')} \Big(1 + \text{FA}(D, x)_{c|j} - \text{FA}(D, x)_{c'|j}  \Big)\\
\end{align*}

Since $d_\text{sym}(D, D') = |D \ominus D'| \geq |h_\text{split} (D\ominus D')|$, we have

\begin{align*}
&\text{FA}(D', x)_c - \text{FA}(D', x)_{c'}\\
&\geq  \text{FA}(D, x)_c - \text{FA}(D, x)_{c'}  \\
&-\frac{1}{k} \max_{H \subseteq [kd] \atop |H|\leq d_\text{sym}(D, D')}
 \sum_{j\in H} \Big(1 + \text{FA}(D, x)_{c|j} - \text{FA}(D, x)_{c'|j}  \Big)\\
&=\text{FA}(D, x)_c - \text{FA}(D, x)_{c'} - \frac{1}{k} \cdot \Delta_{D, x}^{\overline{d_\text{sym} (D, D')}}
\end{align*}

Reorganizing this and subtract $\frac{\mathbb{1}\left[c'<c\right]}{kd}$ from both side, we have
\begin{align*}
&\text{FA}(D', x)_c - \text{FA}(D', x)_{c'} - \frac{\mathbb{1}\left[c'<c\right]}{kd}\\
&\geq \text{FA}(D, x)_c - \text{FA}(D, x)_{c'} - \frac{1}{k} \cdot \Delta_{D, x}^{\overline{d_\text{sym} (D, D')}} - \frac{\mathbb{1}\left[c'<c\right]}{kd}   \\
&\geq 0,
\end{align*}
where the last inequality is from the condition that \begin{align*}
    \frac{1}{k} \cdot \Delta_{D, x}^{\overline{d_\text{sym} (D, D')}}  \leq \text{FA}(D, x)_c - \text{FA}(D, x)_{c'} - \frac{\mathbb{1}\left[c'<c\right]}{kd}
\end{align*}
holds for all $c' \neq c$.

Since $\text{FA}(D', x)_c - \text{FA}(D', x)_{c'} - \frac{\mathbb{1}\left[c'<c\right]}{kd} \geq 0$, we know $\text{FA}(D', x) \neq c'$. Since this holds for any $c' \neq c$, we have $\text{FA}(D', x) = c = \text{FA}(D, x)$, which completes the proof.

\end{proof}

\section{Proof of Theorem \ref{Th:IA}}
\label{Pf:IA}

\begin{proof}

The ideas for this proof are very similar to the ones in proving Theorem \ref{Th:FA}.

Given a training set $D$ and an input $x$, for an arbitrary training set $D'$ such that
\begin{align*}
    \frac{1}{k} \cdot \overline{\Delta}_{D, x}^{\overline{d_\text{sym} (D, D')}}  \leq \text{IA}(D, x)_c - \text{IA}(D, x)_{c'} - \mathbb{1}\left[c'<c\right]\cdot \delta
\end{align*}
holds for all $c' \neq c$ and for some $\delta > 0$, we want to show that $\text{IA}(D', x) = \text{IA}(D, x) = c$.

Let $D \ominus D' = (D\setminus D')\cup (D'\setminus D)$ to denote symmetric difference between $D$ and $D'$, which corresponds to the minimum set of samples to be inserted/removed if one wants change from $D$ to $D'$ (or $D'$ to $D$).

We can reorganize Infinite Aggregation as follows:
\begin{align*}
    \text{IA}(D, X) = &\arg \max_{c \in \mathcal{C}} 
    Pr_{S\sim \text{Bernoulli}(D, \frac{1}{k})}\left[ f_\text{base} (S, x) = c \right]\\
    =&\arg \max_{c \in \mathcal{C}} 
    \mathbb{E}_{S\sim \text{Bernoulli}(D, \frac{1}{k})} \mathbb{1}\left[ f_\text{base} (S, x) = c \right]\\
    =&\arg \max_{c \in \mathcal{C}} 
    \mathbb{E}_{S\sim \text{Bernoulli}(D, \frac{1}{k})} g(S, x)_c,
\end{align*}
where $g(S, x)_c = Pr[f_\text{base}(S,x) = c]$ is the distribution of the predictions and is therefore deterministic.

Since $g(S,x)$ is deterministic, by definition, the contribution from a base classifier will not change if its corresponding training set $S$ remains unchanged from $D$ to $D'$, which is equivalent to $S \subseteq D\cap D'$.

Thus for any $c' \neq c$, we have
\begin{align*}
    &\text{IA}(D', x)_c - \text{IA}(D', x)_{c'}\\
    &=\mathbb{E}_{S \sim Bernoulli(D', \frac{1}{k})}\Big(
        g(S, x)_c - g(S, x)_{c'}
    \Big)\\
    &=\mathbb{E}_{S \sim Bernoulli(D \cup D', \frac{1}{k})}\Big(
        g(S\cap D', x)_c - g(S \cap D', x)_{c'}
    \Big)\\
    &= \text{IA}(D, x)_c - \text{IA}(D, x)_{c'} 
    \\
    &+\mathbb{E}_{S \sim Bernoulli(D \cup D', \frac{1}{k})}
    \mathbb{1}\left[ S \not\subseteq D \cap D' \right] \times 
      \Big( g(S \cap D', x)_c \\
    &- g(S \cap D', x)_{c'}  
    -  g(S \cap D, x)_c + g(S \cap D, x)_{c'}  \Big)\\
    &\geq \text{IA}(D, x)_c - \text{IA}(D, x)_{c'}  
     -\mathbb{E}_{S \sim Bernoulli(D \cup D', \frac{1}{k})} \Big[\\
    &\mathbb{1}\left[ S \not\subseteq D \cap D' \right] \times
    \Big( 1 +  g(S \cap D, x)_c - g(S \cap D, x)_{c'}  \Big) \Big] \\
\end{align*}

Since 
$
    \mathbb{1}\left[ S \not\subseteq D \cap D' \right] \leq 
    \sum_{x_L\in D\ominus D'} \mathbb{1}\left[ x_L\in S \right],
$
we can further bound the above formula as follows:
\begin{align*}
    &\text{IA}(D', x)_c - \text{IA}(D', x)_{c'}\\
    &\geq \text{IA}(D, x)_c - \text{IA}(D, x)_{c'}  
     -\mathbb{E}_{S \sim Bernoulli(D \cup D', \frac{1}{k})} \Big[\\
    &\sum_{x_L\in D\ominus D'} \mathbb{1}\left[ x_L\in S \right] \times
    \Big( 1 +  g(S \cap D, x)_c - g(S \cap D, x)_{c'}  \Big) \Big] \\
    &= \text{IA}(D, x)_c - \text{IA}(D, x)_{c'}  \\
    & -\sum_{x_L\in D\ominus D'} \mathbb{E}_{S \sim Bernoulli(D \cup D', \frac{1}{k})} \Big[\\
    &\mathbb{1}\left[ x_L\in S \right] \times
    \Big( 1 +  g(S \cap D, x)_c - g(S \cap D, x)_{c'}  \Big) \Big] \\
    &= \text{IA}(D, x)_c - \text{IA}(D, x)_{c'} \\
    & -\sum_{x_L\in D\setminus D'} \frac{1}{k}\Big( 1 + \text{IA}(D, x)_{c|x_L} - \text{IA}(D, x)_{c'|x_L} \Big) \\
    &- \sum_{x_L\in D'\setminus D} \frac{1}{k} \Big( 1 + \text{IA}(D, x)_c - \text{IA}(D, x)_{c'} \Big) \\
    &\geq \text{IA}(D, x)_c - \text{IA}(D, x)_{c'} 
    - \frac{1}{k}\overline{\Delta}_{D, x}^{\overline{d_\text{sym} (D, D')}}
\end{align*}

Subtracting both sides with $\mathbb{1}\left[c'<c\right]\cdot \delta$,
we have
\begin{align*}
&\text{IA}(D', x)_c - \text{IA}(D', x)_{c'} - \mathbb{1}\left[c'<c\right]\cdot \delta \\
&\geq \text{IA}(D, x)_c - \text{IA}(D, x)_{c'} - \mathbb{1}\left[c'<c\right]\cdot \delta
    - \frac{1}{k}\overline{\Delta}_{D, x}^{\overline{d_\text{sym} (D, D')}} \\
&\geq 0
\end{align*}
for some $\delta > 0$, where the last inequality is from the condition that 
\begin{align*}
    \frac{1}{k} \cdot \overline{\Delta}_{D, x}^{\overline{d_\text{sym} (D, D')}}  \leq \text{IA}(D, x)_c - \text{IA}(D, x)_{c'} - \mathbb{1}\left[c'<c\right]\cdot \delta
\end{align*}
holds for all $c' \neq c$ and for some $\delta > 0$.

This means that $\text{IA}(D', x) \neq c'$. 
Since this is true for any $c' \neq c$, we have
$\text{IA}(D', x) = c = \text{IA}(D, x)$, 
which completes the proof.

\end{proof}

\section{Certified Accuracy from Finite Aggregation}
\label{appendix:FA_cert_acc}
In this section, we show how certified accuracy can be computed for Finite Aggregation (i.e. how to derive a lower bound of accuracy on a given test set under poisoning attacks). The key in this derivation is to realize that using {\it same} poisons for multiple targets meaning a more restricted subset of partitions affected compared to different poisons.

We introduce a variant of Theorem \ref{Th:FA} as follows, certifying a prediction under conditional poisoning attacks, where only a given subset of partitions (i.e. $Q \subseteq [kd]$) may be affected by poisons.

\begin{myTh}[Certificates under Conditional Poisoning]
\label{Th:FA_cert_acc}
Given a training set $D$, a subset of partition indices $Q \subseteq [kd]$, and an input $x$, let $c = \text{FA}(D, x)$, then for any training set $D'$ such that $h_\text{split}\left( (D\setminus D')\cup (D'\setminus D) \right)\subseteq Q$, it is guaranteed that $\text{FA}(D', x) = \text{FA}(D, x)$ when
\begin{align*}
    \frac{1}{k} \cdot \Delta_{D, Q, x}^{\overline{d_\text{sym} (D, D')}}  \leq \text{FA}(D, x)_c - \text{FA}(D, x)_{c'} - \frac{\mathbb{1}\left[c'<c\right]}{kd}
\end{align*}
holds for all $c' \neq c$, where $\Delta_{D, Q, x}$ is a multiset defined as 
\begin{align*}
\left\{ 1 + \text{FA}(D, x)_{c|j} - \text{FA}(D, x)_{c'|j} \right\}_{j \in Q}
\end{align*}
and $\Delta_{D, Q, x}^{\overline{d_\text{sym} (D, D')}}$ denotes the sum of the largest $d_\text{sym} (D, D')$ elements in the multiset $\Delta_{D, Q, x}$.
\end{myTh}

The proof is in Appendix \ref{Pf:FA_cert_acc}. 
Theorem \ref{Th:FA_cert_acc} provides a sufficient condition for when the prediction on this sample from Finite Aggregation is certifiably correct under poisoning attacks.
Given a training set $D$, a subset of partitions $Q\subseteq [kd]$, an attack budget $d_\text{sym}$ (i.e. number of poisons allowed), for any test sample with input $x$ and ground truth label $c$, we define $\mathcal{O}_{x, c, Q, d_\text{sym}}$ to be $1$ when the prediction is certifiably correct under poisoning attacks and $0$ otherwise. 

Consequently, the certified accuracy on the test set $D_\text{test}$ with an attack budget $d_\text{sym}$ can be expressed as follows by definition:
\begin{align*}
    \min_{Q \subseteq [kd] \atop |Q| \leq d_\text{sym}}
    \frac{1}{|D_\text{test}|} 
    \sum_{(x, c)\in D_\text{test}} \mathcal{O}_{x, c, Q, d_\text{sym}}.
\end{align*}

\section{Proof of Theorem \ref{Th:FA_cert_acc}}
\label{Pf:FA_cert_acc}
\begin{proof}

Given a training set $D$, a subset of partition indices $Q \subseteq [kd]$, and an input $x$, for an arbitrary training set $D'$ such that
$h_\text{split}\left( (D\setminus D')\cup (D'\setminus D) \right)\subseteq Q$ and 
\begin{align*}
    \frac{1}{k} \cdot \Delta_{D, Q, x}^{\overline{d_\text{sym} (D, D')}}  \leq \text{FA}(D, x)_c - \text{FA}(D, x)_{c'} - \frac{\mathbb{1}\left[c'<c\right]}{kd}
\end{align*}
holds for all $c' \neq c$, we want to show that $\text{FA}(D', x) = \text{FA}(D, x) = c$.

Let $D \ominus D' = (D\setminus D')\cup (D'\setminus D)$ to denote symmetric difference between $D$ and $D'$, which corresponds to the minimum set of samples to be inserted/removed if one wants change from $D$ to $D'$ (or $D'$ to $D$).

Since the base classification algorithm $f_\text{base}$ is deterministic, by definition, the prediction from a base classifier will not change if its corresponding training set $S_i$ ($S_i'$) remains unchanged from $D$ to $D'$, which is equivalent to $i \notin \bigcup_{x \in D\ominus D'} h_\text{spread}(h_\text{split(x)})$.
Let $$h_{D\ominus D'} = \bigcup_{x \in D\ominus D'} h_\text{spread}(h_\text{split(x)}).$$ 

Thus for any $c' \neq c$, we have
\begin{align*}
    &\text{FA}(D', x)_c - \text{FA}(D', x)_{c'}\\
    &=\frac{1}{kd} \sum_{i\in [kd]}\Big( \mathbb{1}\left[ 
    f_\text{base}(S'_i, x)=c\right] - \mathbb{1}\left[ 
    f_\text{base}(S'_i, x)=c'\right] \Big)\\
    &= \text{FA}(D, x)_c - \text{FA}(D, x)_{c'}  \\
    &+\frac{1}{kd} 
     \sum_{i\in h_{D\ominus D'}}\Big( \mathbb{1}\left[ 
    f_\text{base}(S'_i, x)=c\right] - \mathbb{1}\left[ 
    f_\text{base}(S'_i, x)=c'\right]  \Big)  \\
    &-\frac{1}{kd} 
     \sum_{i\in h_{D\ominus D'}} \Big( \mathbb{1}\left[ 
    f_\text{base}(S_i, x)=c\right] - \mathbb{1}\left[ 
    f_\text{base}(S_i, x)=c'\right]  \Big)\\
    &\geq \text{FA}(D, x)_c - \text{FA}(D, x)_{c'}  \\
    &+\frac{1}{kd} 
     \sum_{i\in h_{D\ominus D'}} \Big( 0 - 1  \Big)  \\
    &-\frac{1}{kd} 
    \sum_{i\in h_{D\ominus D'}} \Big( \mathbb{1}\left[ 
    f_\text{base}(S_i, x)=c\right] - \mathbb{1}\left[ 
    f_\text{base}(S_i, x)=c'\right]  \Big)\\
    & = \text{FA}(D, x)_c - \text{FA}(D, x)_{c'}  -\\
    &\frac{1}{kd} 
     \sum_{i\in h_{D\ominus D'}} \Big(1 + \mathbb{1}\left[ 
    f_\text{base}(S_i, x)=c\right] - \mathbb{1}\left[ 
    f_\text{base}(S_i, x)=c'\right]  \Big)\\
\end{align*}

We can rewrite $h_{D\ominus D'}$ as follows:
\begin{align*}
h_{D\ominus D'} = \bigcup_{x \in D\ominus D'} h_\text{spread}(h_\text{split(x)}) = \bigcup_{j \in h_\text{split}(D\ominus D')} h_\text{spread}(j),
\end{align*}
where $h_\text{split}(D\ominus D') = \{ h_\text{split}(x) | x\in D\ominus D' \}$.

Since $1 + \mathbb{1}\left[ f_\text{base}(S_i, x)=c\right] - \mathbb{1}\left[ f_\text{base}(S_i, x)=c'\right] \geq 0$,
we can further bound the above formula as follows:
\begin{align*}
&\text{FA}(D', x)_c - \text{FA}(D', x)_{c'}\\
&\geq  \text{FA}(D, x)_c - \text{FA}(D, x)_{c'}  \\
&-\frac{1}{kd} 
 \sum_{j\in h_\text{split}(D \ominus D')}\Big( \sum_{i \in h_\text{spread}(j)} 1 + \mathbb{1}\left[ 
f_\text{base}(S_i, x)=c\right]  \\
& - \mathbb{1}\left[ 
f_\text{base}(S_i, x)=c'\right]  \Big)\\
&=\text{FA}(D, x)_c - \text{FA}(D, x)_{c'}  \\
&-\frac{1}{k} 
 \sum_{j\in h_\text{split}(D \ominus D')} \Big(1 + \text{FA}(D, x)_{c|j} - \text{FA}(D, x)_{c'|j}  \Big)\\
\end{align*}

Since $d_\text{sym}(D, D') = |D \ominus D'| \geq |h_\text{split} (D\ominus D')|$ and $h_\text{split} (D\ominus D') \subseteq Q$, we have

\begin{align*}
&\text{FA}(D', x)_c - \text{FA}(D', x)_{c'}\\
&\geq  \text{FA}(D, x)_c - \text{FA}(D, x)_{c'}  \\
&-\frac{1}{k} \max_{H \subseteq Q \atop |H|\leq d_\text{sym}(D, D')}
 \sum_{j\in H} \Big(1 + \text{FA}(D, x)_{c|j} - \text{FA}(D, x)_{c'|j}  \Big)\\
&=\text{FA}(D, x)_c - \text{FA}(D, x)_{c'} - \frac{1}{k} \cdot \Delta_{D, Q, x}^{\overline{d_\text{sym} (D, D')}}
\end{align*}

Reorganizing this and subtract $\frac{\mathbb{1}\left[c'<c\right]}{kd}$ from both side, we have
\begin{align*}
&\text{FA}(D', x)_c - \text{FA}(D', x)_{c'} - \frac{\mathbb{1}\left[c'<c\right]}{kd}\\
&\geq \text{FA}(D, x)_c - \text{FA}(D, x)_{c'} - \frac{1}{k} \cdot \Delta_{D, Q, x}^{\overline{d_\text{sym} (D, D')}} - \frac{\mathbb{1}\left[c'<c\right]}{kd}   \\
&\geq 0,
\end{align*}
where the last inequality is from the condition that \begin{align*}
    \frac{1}{k} \cdot \Delta_{D, Q, x}^{\overline{d_\text{sym} (D, D')}}  \leq \text{FA}(D, x)_c - \text{FA}(D, x)_{c'} - \frac{\mathbb{1}\left[c'<c\right]}{kd}
\end{align*}
holds for all $c' \neq c$.

Since $\text{FA}(D', x)_c - \text{FA}(D', x)_{c'} - \frac{\mathbb{1}\left[c'<c\right]}{kd} \geq 0$, we know $\text{FA}(D', x) \neq c'$. Since this holds for any $c' \neq c$, we have $\text{FA}(D', x) = c = \text{FA}(D, x)$, which completes the proof.

\end{proof}

\end{document}